\documentclass[letterpaper,10pt,conference]{ieeeconf} 
\IEEEoverridecommandlockouts       
\usepackage{graphics} 
\usepackage{epsfig} 
\usepackage[]{algorithmic}
\usepackage[]{algorithm}
\usepackage{amsmath,amssymb,amsopn,amstext,amsfonts}
\usepackage{color}
\usepackage{multirow}
\usepackage{subfigure}
\usepackage{makecell}
\usepackage{hyperref}

\DeclareGraphicsExtensions{.pdf,.png,.jpg}

\title{\LARGE \bf Probabilistic Dense Reconstruction from a Moving Camera}

\author{Yonggen Ling$^1$, Kaixuan Wang$^2$, and Shaojie Shen$^2$
\thanks{$^1$Tencent AI Lab, China. $^2$The Hong Kong University of Science and Technology, Hong Kong, SAR China.  Correspondence to: Yonggen Ling {\tt\small ylingaa@connect.ust.hk}, Kaixuan Wang and Shaojie Shen {\tt\small \{kwangap, eeshaojie\}@ust.hk}. This work was partially supported by  HKUST institutional studentship.}
}
\hypersetup{draft} 
\begin{document}

\maketitle
\thispagestyle{empty}
\pagestyle{empty}

\begin{abstract}
This paper presents a probabilistic approach for online dense reconstruction using a single monocular camera moving through the environment. Compared to spatial stereo, depth estimation from motion stereo is challenging due to insufficient parallaxes, visual scale changes, pose errors, etc. We utilize both the spatial and temporal correlations of consecutive depth estimates to increase the robustness and accuracy of monocular depth estimation. An online, recursive, probabilistic scheme to compute depth estimates, with corresponding covariances and inlier probability expectations, is proposed in this work. We integrate the obtained depth hypotheses into dense 3D models in an uncertainty-aware way. We show the effectiveness and efficiency of our proposed approach by comparing it with state-of-the-art methods in the TUM RGB-D SLAM \& ICL-NUIM dataset. Online indoor and outdoor experiments are also presented for performance demonstration.
\end{abstract}

\section{Introduction}
Accurate localization and dense mapping are fundamental components of autonomous robotic systems as they serve as the perception input for obstacle avoidance and path planning. While localization from a monocular camera has been well discussed in the past \cite{SheMicKum1505,HesKotBow1402,LiMou1305,orb-slam,forster15}, online dense reconstruction using a single moving camera is still under development \cite{NewcombeLD11,REMODE,MonoFusion,alpha_beta}. Since monocular depth estimation is based on consecutive estimated poses and images, main issues of it are: imprecise poses due to localization errors, inaccurate visual correspondences due to insufficient parallaxes and visual scale changes, etc. Depth estimation from traditional spatial stereo cameras (usually in the front-parallel setting), however, avoids the issues met with motion stereo. Thus many algorithms based on stereo cameras have been developed in the past decades \cite{sgm07,Geiger2010ACCV}. The significant drawback of spatial stereo is its baseline limitation: distant objects can be better estimated using longer baselines because of larger disparities; while close-up structures can be better reconstructed using shorter baselines because of larger visual overlaps. Moreover, for real world applications such as mobile robots, phones and wearable devices, it is impossible to equip them with long baseline stereo cameras because of the size constraint. If the baseline length, compared to the average scene depth of the perceived environment, is relatively small, images captured on stereo cameras will be similar. As a result, visual information from stereo cameras degrades to the same level as that obtained by a monocular camera. 

\begin{figure}[!h]
	\centering
	\subfigure[Dense indoor reconstruction for motion planning.]{\includegraphics[width=0.48\columnwidth]{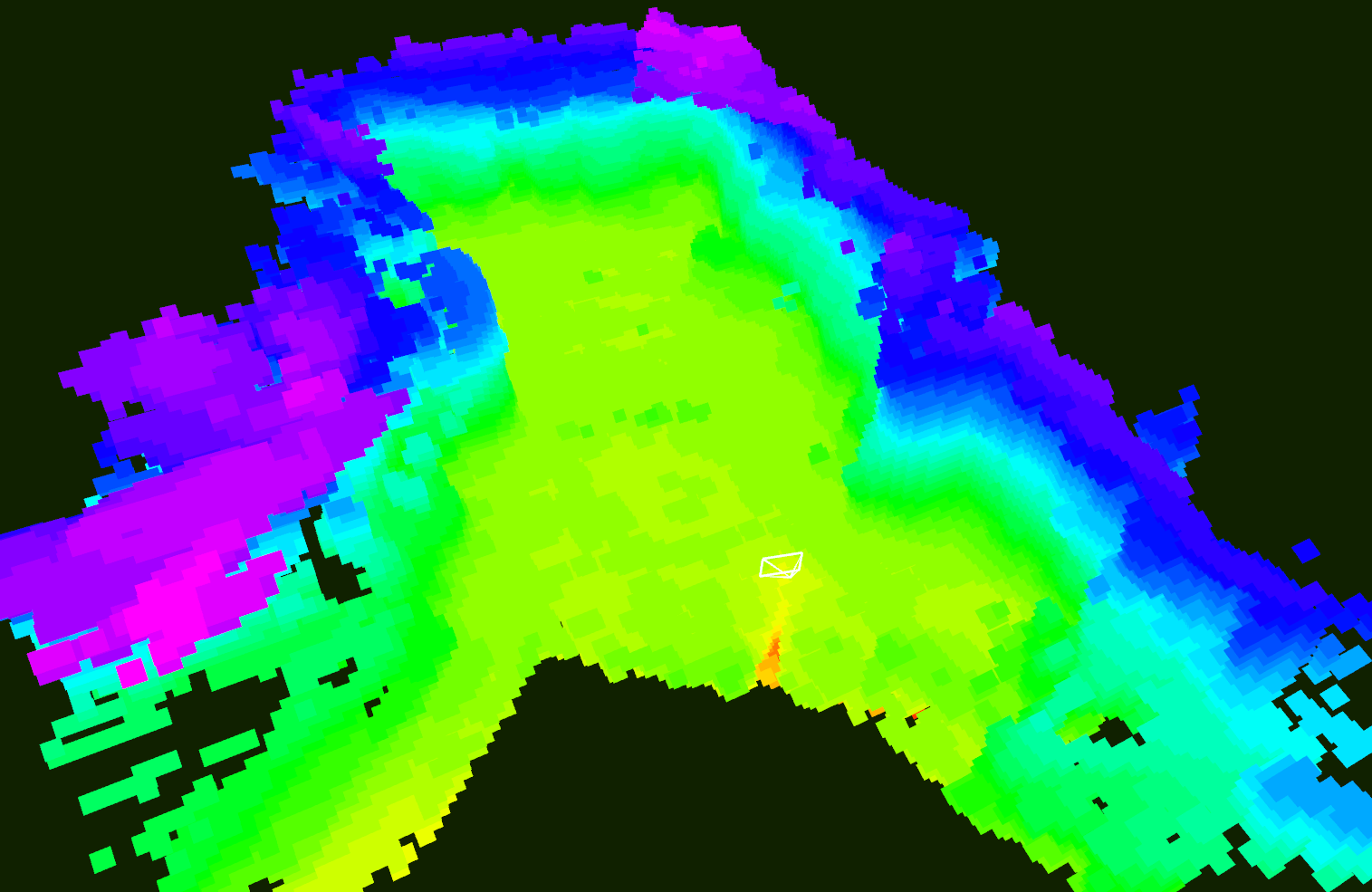}}
	\subfigure[Meshing view of indoor reconstruction for visualization.]{\includegraphics[width=0.48\columnwidth]{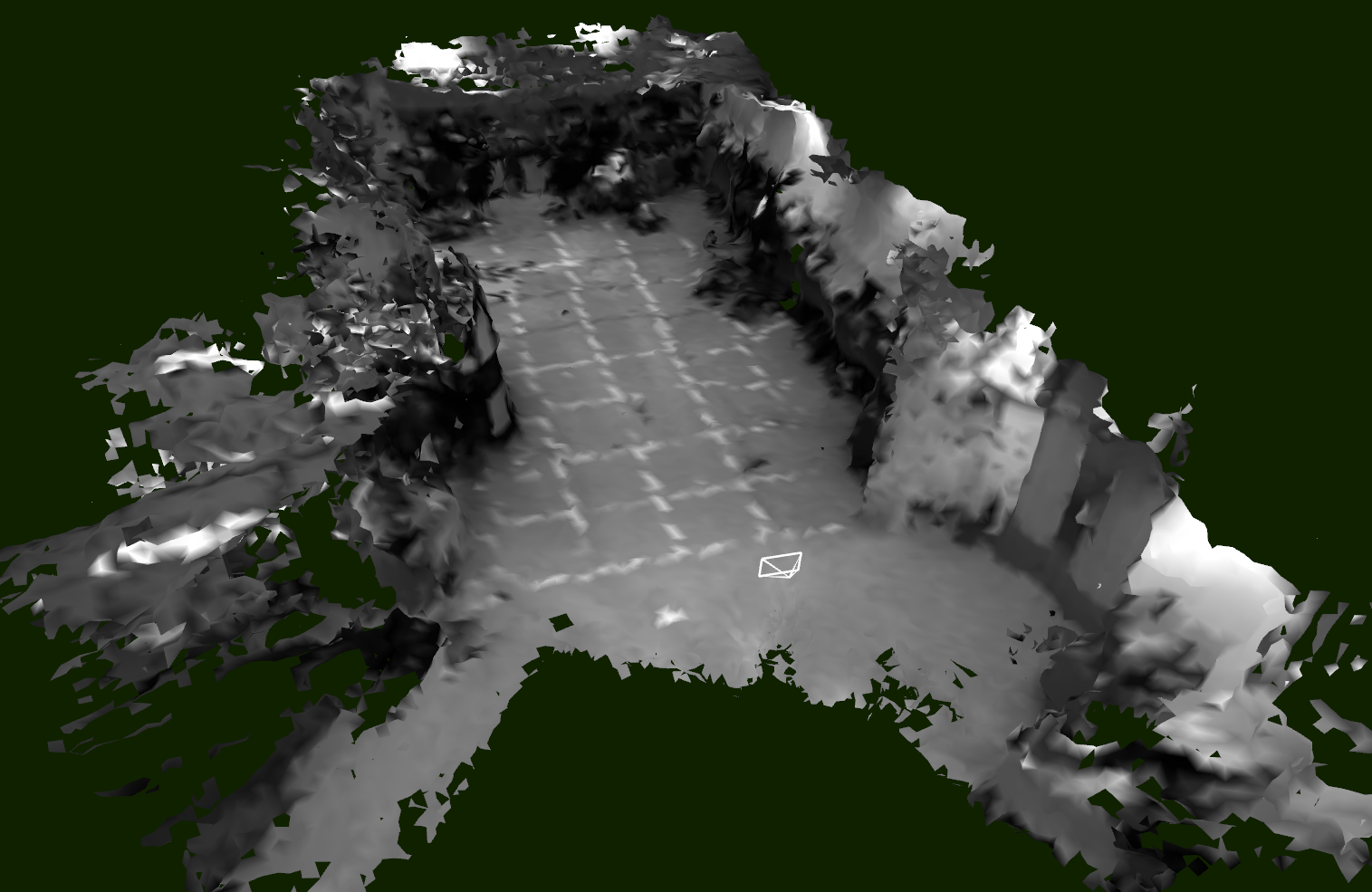}}
	\subfigure[Dense outdoor reconstruction for motion planning.]{\includegraphics[width=0.48\columnwidth]{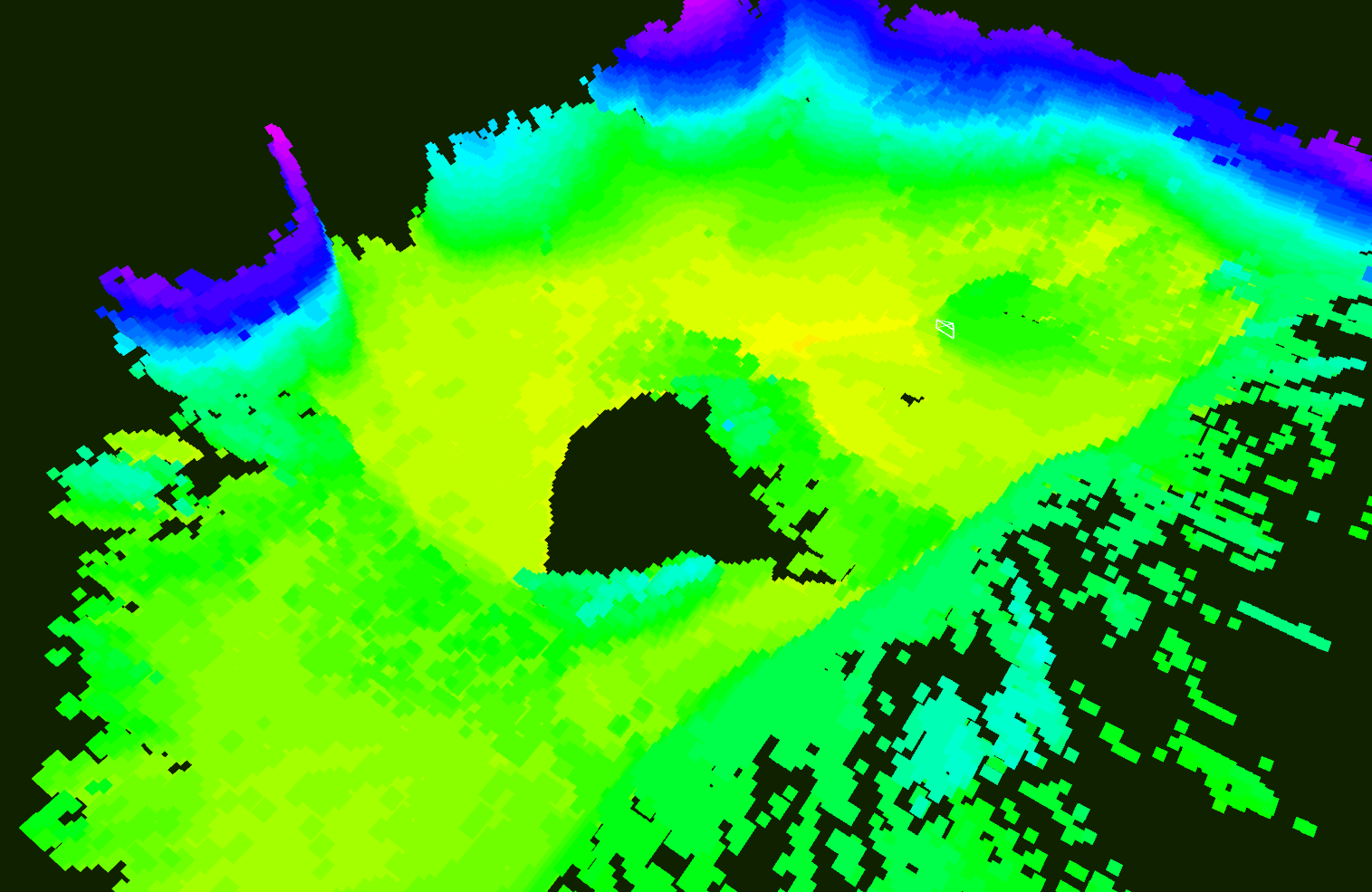}}
	\subfigure[Meshing view of indoor reconstruction for visualization.]{\includegraphics[width=0.48\columnwidth]{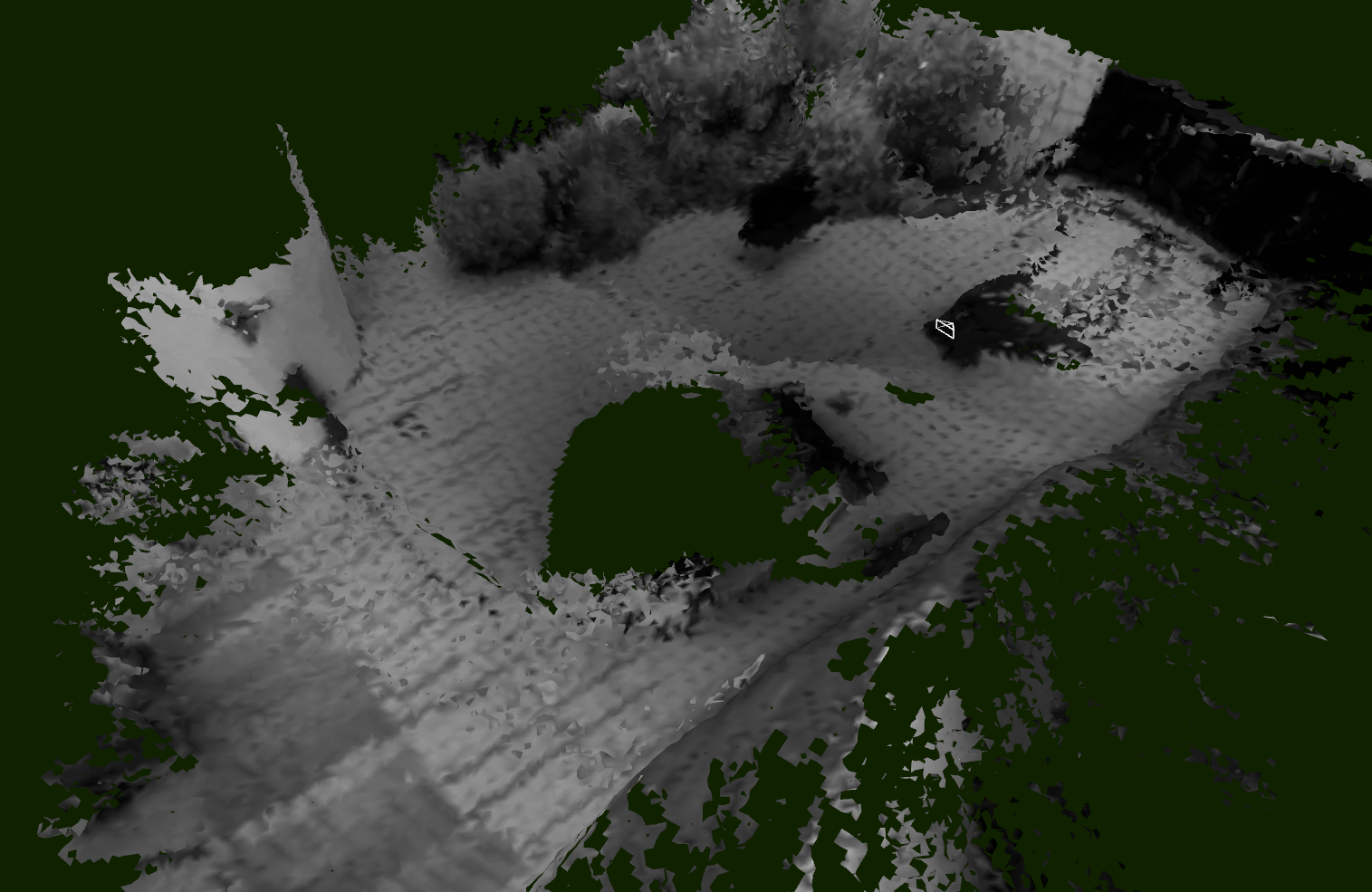}}
	\caption[Dense reconstruction of an indoor/outdoor environment from a single moving camera.]{Dense reconstruction of an indoor/outdoor environment from a single moving camera. (a)(c) Reconstruction for robotic applications, such as motion planning and obstacle avoidance. Colors vary w.r.t. the height to show the structure of the reconstructed dense environment. (b)(d) Meshing view by applying marching cubes \cite{marching_cubes} on TSDFs for visualization. More details can be found at: \url{https://1drv.ms/v/s!ApzRxvwAxXqQmlW9ZOrp9hdA7ude}. }
	\label{fig:exp_indoor.}
\end{figure}

Fundamentally different from passive cameras, time-of-flight (TOF) cameras as well as structure-light cameras, emit light actively. They are able to provide high accuracy depth measurements. With the advent of Microsoft Kinect and ASUS Xtion, dense reconstruction algorithms based on active depth cameras \cite{kinectfusion,Niebner2013,Whelan16ijrr} have achieved impressive results in recent years. Unfortunately, active sensors do not work under strong sunlight, which limits their application to indoor environments.

This paper focuses on dense reconstructions using a single monocular camera, which adapts to both indoor and outdoor environments with various scene depth ranges. Comparing to existing methods \cite{vi_mean,alpha_beta,REMODE,kinectfusion,Niebner2013,Whelan16ijrr}, we make careful improvements to multiple sub-modules of the whole mapping pipeline, resulting in substantial gains in the mapping performance. The main contributions of this paper are as follows:
\begin{itemize}
	\item A joint probabilistic consideration of depth estimation and integration.
	\item A detailed discussion of aggregated costs and their probability modeling.
	\item An online, recursive, probabilistic depth estimation scheme that utilizes both the spatial and temporal correlations of consecutive depth estimates.
	\item Open-source implementations available at \url{https://github.com/ygling2008/probabilistic_mapping}.
\end{itemize}
To validate the effectiveness and efficiency of the proposed approach, we compare it with state-of-the-art methods on the TUM RGB-D SLAM \& ICL-NUIM dataset. We also demonstrate its online performance on indoor and outdoor dense reconstructions. 

The rest of this paper is structured as follows. Sect.~\ref{sec:related_work} reviews the related work. Our proposed approach is presented in Sect.~\ref{sec:probabilistic_mapping}, with experimental comparisons and validations demonstrated in Sect.~\ref{sec:experiments}. Sect~\ref{sec:conclusions} draws the conclusion and points out possible future extensions.

\section{Related Work}
\label{sec:related_work} 
There has been extensive scholarly work on reconstructing a scene from images collected by a single moving camera. We only discuss the works most related to ours, that is, online monocular dense reconstruction systems.

Early live dense reconstruction systems are proposed by Stuhmer1 et al.~\cite{Stuhmer2010} and Newcombe et al.~\cite{NewcombeLD11}, where the problem of dense reconstruction is formulated as an optimization problem. They solve for all depth values in multiple views by jointly minimizing the intensity difference and depth discontinuity. While Stuhmer1 et al.~\cite{Stuhmer2010} rely on feature tracking for localization, Newcombe et al.~\cite{NewcombeLD11} use the built dense reconstruction for pose tracking.  Optimization-based methods are computationally intensive, thus they are usually run on high-performance GPUs.

To resolve the demanding computations, \cite{alpha_beta} ignores the spatial correlation between neighboring depth estimates, and computes each depth independently. \cite{MonoFusion}, \cite{3DModelingOnTheGO}, and \cite{REMODE} decouple the constraints of photometric consistency and depth continuity. They firstly search for the optimal depth estimate for every pixel and then regularize the computed depths to enforce the consistency between neighboring depth estimates. Various filters are also included for outlier detection and removal. While these relaxations greatly reduce the algorithmic complexity, mapping results of these approaches are not as good as those of the optimization-based methods.
Another relaxation is to narrow the depth searching range by merely evaluating depth values within a limited number of discrete depth samples \cite{vi_mean}. \cite{vi_mean} uses the dynamic programming scheme proposed in semi-global matching (SGM) \cite{sgm07} for cost minimization. It runs fast; however, its depth estimation contains many outliers as it neither makes use of the temporal correlation in image sequences nor deals with outliers.

The last algorithms to mention are those that reconstruct dense 3D models from sparse features or semi-dense mapping results \cite{multi_level_mapping,dpptam,superpixel_expansion}. \cite{multi_level_mapping} computes depth in multiple levels of images and then combines the obtained results into a final one. The density of mapping outputs from \cite{multi_level_mapping} depends on environments that are suitable for multi-level matching. Based on the local planar assumption, \cite{dpptam} and \cite{superpixel_expansion} use superpixels to expand the built semi-dense maps to dense mappings. \cite{dpptam} and \cite{superpixel_expansion} run fast; however, their superpixel extraction algorithms are not robust, with many ambiguities for superpixel segmentation. The effectiveness of the local planar assumption depends on the quality of superpixel extraction.

The most similar work to ours is \cite{REMODE}. While \cite{REMODE} adopts total variation smoothing for incorporating depth continuity, our work uses the dynamic programming scheme \cite{sgm07} instead. Moreover, we utilize both the spatial and temporal correlations inherent to image sequences in the whole probabilistic and recursive depth estimation process. \cite{REMODE} decouples them into two separate steps. We consider more cases of cost aggregation and probability modeling than \cite{REMODE}, and we also introduce an uncertainty-aware depth integration for dense reconstruction, which is not covered in \cite{REMODE}.

\section{Monocular Dense Reconstruction}
\label{sec:probabilistic_mapping}
Our reconstruction pipeline is shown in Fig.~\ref{fig:gaussian_beta}. It consists of three steps: depth estimation, hypothesis filtering, and uncertainty-aware depth integration.

\subsection{Depth Estimation via Motion Stereo}\
\label{subsec:depth_estiimation}
Our dense reconstruction system is built upon a feature-based SLAM pipeline, which provides camera poses in real time. This SLAM pipeline can be vision-based \cite{orb-slam} or visual-inertial-based \cite{HesKotBow1402,forster15,zhenfei_tase}. For each incoming keyframe image, we compute its corresponding depth estimation.

\subsubsection{Temporal Cost Aggregation}
\label{subsubsec:T}
We set the latest incoming keyframe as the reference frame, and aggregate information from past frames. $K_a$ ($K_a=5$) frames spanning various parallax ranges are selected. They uniformly cover the average parallax deviation, ranging from 0 to $K_p$ ($K_p$ = 100) pixels, from the reference frame. This deviation is computed as the average corner location difference of the tracked features with rotation compensation. We select past frames based on the parallax deviation instead of the actual distance for adaption to environments with various scene depths.

For the benefit of online computation, we restrict every depth estimate to be one of $L$ ($L$ = 64) depth samples. These $L$ depth samples are not uniformly distributed within the feasible depth range. Instead they follow the principle of \textit{depth from disparity}: each depth $d$ is a function of its disparity $disp$, baseline length $b$ and focus length $f$,
\begin{flalign}
d = \frac{bf}{disp} = \frac{1}{disp \cdot \frac{1}{bf}} = \frac{1}{disp \cdot c_d}
\end{flalign}
where $c_d = \frac{1}{bf}$. Baseline length $b$ is set depending on the average depth of the perceived environment. We enumerate $disp$ from 0 to $L-1$, and obtain the set of $L$ depth samples $\Phi(L) = \{  \frac{1}{63 \cdot c_d}, \frac{1}{62 \cdot c_d}, ..., \infty \}$.
Given a pixel $\mathbf{u}_i$ in the reference image $i$ as well as its depth $d_\mathbf{u} \in \Phi(L)$, we project it on its aggregation frame $j \in K_a$ with pixel coordinate $\mathbf{u}_j$:
\begin{flalign}
\begin{bmatrix} 
\mathbf{u}_j \\ 
1
\end{bmatrix} &\simeq \mathbf{K} \mathbf{R}_w^j (\mathbf{R}_i^w d_\mathbf{u} \mathbf{K}^{-1}\begin{bmatrix} 
\mathbf{u}_i \\ 
1
\end{bmatrix} + \mathbf{t}_i^w - \mathbf{t}_j^w ) = d_\mathbf{u} \mathbf{h}_{i}^j + \mathbf{c}_{i}^j
\end{flalign}
where $\mathbf{h}_{i}^j =  \mathbf{K} \mathbf{R}_w^j \mathbf{R}_i^w \mathbf{K}^{-1}\begin{bmatrix} 
\mathbf{u}_i \\ 
1
\end{bmatrix} $, $\mathbf{c}_{i}^j = \mathbf{K} \mathbf{R}_w^j(\mathbf{t}_i^w - \mathbf{t}_j^w)$, $\mathbf{K}$ is the camera matrix, and $\mathbf{R}_i^w$, $\mathbf{R}_j^w$ and $\mathbf{t}_i^w$, $\mathbf{t}_j^w$ are rotations and translations of images $i$ and $j$ w.r.t. the world frame respectively. The cost $e(\mathbf{u}_i, d_\mathbf{u}, \mathbf{u}_j)$ between $\mathbf{u}_i$ and $\mathbf{u}_j$ given $d_\mathbf{u}$ is the sum of the absolute differences between intensities within two $3\times 3$ patches centered on $\mathbf{u}_i$ and $\mathbf{u}_j$. We define the cost of pixel $\mathbf{u}_i$ with depth estimate $d_\mathbf{u}$ as $ e(\mathbf{u}_i, d_\mathbf{u}) $, which is the aggregation of costs $e(\mathbf{u}_i, d_\mathbf{u}, \mathbf{u}_j)$ from $K_a$ selected frames:
\begin{flalign}
 e(\mathbf{u}_i, d_\mathbf{u}) = \frac{1}{N_a} \sum_{j \in N_a}  e(\mathbf{u}_i, d_\mathbf{u}, \mathbf{u}_j)
\end{flalign}
where $N_a$ ($ <= K_a$) is the number of $\mathbf{u}_j$ within the image size after projection.

\subsubsection{Spatial Regulation}
\label{subsubsec:S}
We notice that using a simple winner-takes-all strategy after the cost aggregation step does not produce reliable depth estimate, as it does not capture the piece-wise linear nature of depth images. In addition, in regions that are texture-less or with repetitive pattern, aggregated cost at a branch of depths are similar. As a result, the depth estimate from the winner-takes-all strategy is greatly affected by the image noise. We thus incorporate the spatial constraints between neighboring depths by using the semi-global optimization proposed in \cite{sgm07}. The 4-path dynamic programming is adopted for the balance between complexity and accuracy.

\begin{figure}[h]
	\centering
	\subfigure[]{\includegraphics[width=0.4\columnwidth]{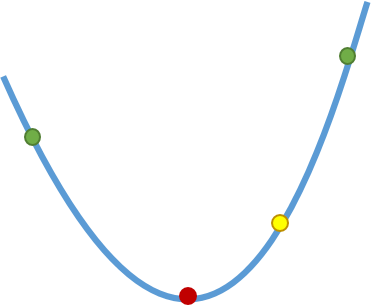}}  \ \ \ \
	\subfigure[]{\includegraphics[width=0.4\columnwidth]{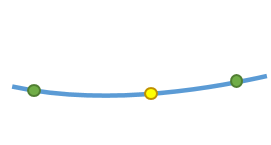}} 
	\caption[The local region around the optimal depth value (shown in yellow).]{The local region around the optimal depth estimate (shown in yellow): (a) NOT flat, (b) flat. The previous and next depth sample of the optimal depth estimate are shown in green. Refined depths are shown in red.}
	\label{fig:p_cases}
\end{figure}

\subsubsection{Local Region Discussion \& Depth Refinement}
\label{subsubsec:D}
We define $S(\mathbf{u}_i, d_\mathbf{u}) $ as the cost of pixel $\mathbf{u}_i$ with depth $d_\mathbf{u}$ after the 4-path aggregation \cite{sgm07} in the previous step. We can take $d_\mathbf{u}^* = \min_{d_\mathbf{u}} S(\mathbf{u}_i, d_\mathbf{u})$ as the output depth estimate. However, since $d_\mathbf{u}$ is one of discrete samples from $\Phi(L)$, its accuracy may not be high. We are going to refine the output depth estimate. 
We examine the local region around the optimal depth value $d_\mathbf{u}^* = \min_{d_\mathbf{u}} S(\mathbf{u}_i, d_\mathbf{u})$ (i.e. the yellow point in Fig.~\ref{fig:p_cases}). Let $d_\mathbf{u}^{*-}$ and $d_\mathbf{u}^{*+}$ be the previous and next depth sample of $d_\mathbf{u}^*$ respectively (i.e. the green point in Fig.~\ref{fig:p_cases}). There are two cases, as shown in Fig.~\ref{fig:p_cases}. In case (a), the local region around the optimal depth value $d_\mathbf{u}^* $ is NOT flat, and we use parabola interpolation to improve the depth estimate accuracy:
\begin{flalign}
S(\mathbf{u}_i, d_\mathbf{u}^{*-}) &= c_0{d_\mathbf{u}^{*-}}^2 + c_1d_\mathbf{u}^{*-} + c_2 \\
S(\mathbf{u}_i, d_\mathbf{u}^{*}) &= c_0{d_\mathbf{u}^{*}}^2 + c_1d_\mathbf{u}^{*} + c_2  \\
S(\mathbf{u}_i, d_\mathbf{u}^{*+}) &= c_0{d_\mathbf{u}^{*+}}^2 + c_1d_\mathbf{u}^{*+} + c_2  
\end{flalign}
where $c_0$, $c_1$, and $c_2$ are three parabola parameters. Solving the above equations, we get the refined depth estimate (i.e. the red point in Fig.~\ref{fig:p_cases}~(a)):
\begin{flalign}
d_\mathbf{u}^{*} \leftarrow d_\mathbf{u}^{*} - \frac{1}{2} \frac{S(\mathbf{u}_i, d_\mathbf{u}^{*+})-S(\mathbf{u}_i, d_\mathbf{u}^{*-})}{S(\mathbf{u}_i, d_\mathbf{u}^{*+}) + S(\mathbf{u}_i, d_\mathbf{u}^{*-}) - 2S(\mathbf{u}_i, d_\mathbf{u}^{*}) }.
\end{flalign}
In case (b), where the local region around the optimal depth value $d_\mathbf{u}^* $ is flat, i.e., $2 \times (1+\epsilon_d) \times S(\mathbf{u}_i, d_\mathbf{u}^{*}) > S(\mathbf{u}_i, d_\mathbf{u}^{*-}) + S(\mathbf{u}_i, d_\mathbf{u}^{*+})$ and $\epsilon_d=0.05$ ($\epsilon_d$ can also be learned using opened RGB-D datasets), depth estimation is not reliable. We regard this depth estimate as an outlier depth estimate.

Note that different cases of local regions result in different probability modelings and update schemes (Sect.~\ref{subsec:probabilistic_filtering}).


\subsection{Hypothesis Filtering via Bayesian Gaussian Beta Process}
\label{subsec:probabilistic_filtering}
\begin{figure*}[!t]
	\centering
	\includegraphics[width=2.0\columnwidth]{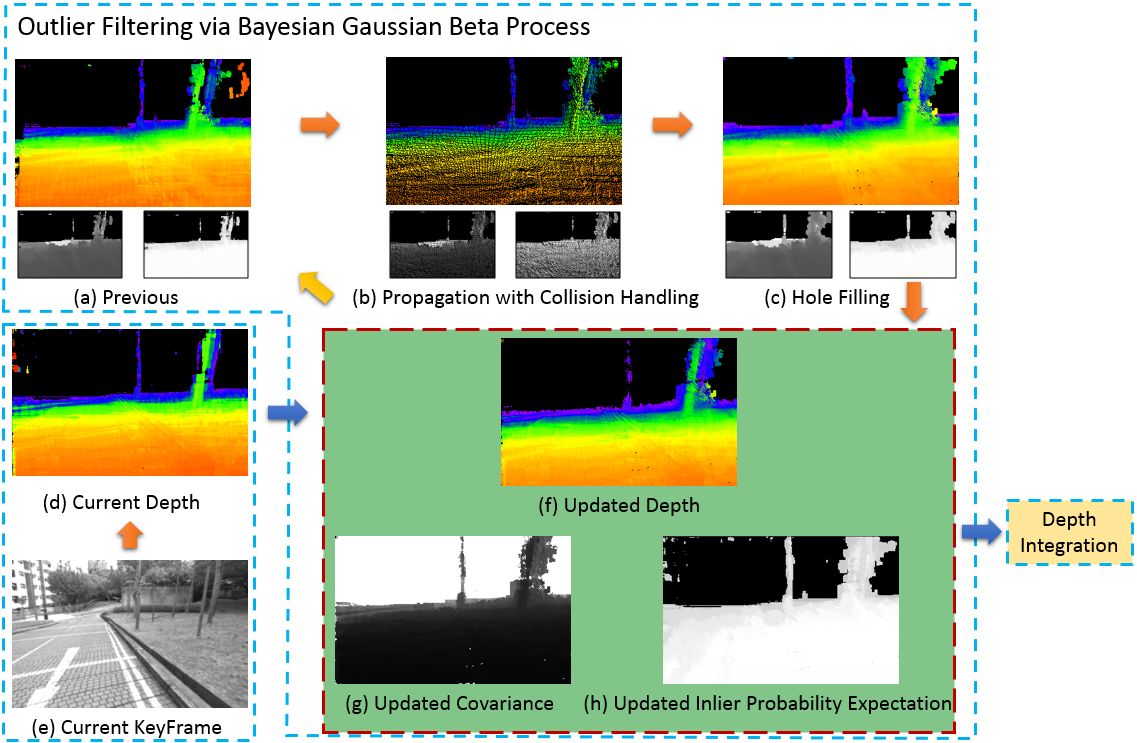}
	\caption[An illustration of the Bayesian Gaussian beta process.]{An illustration of the Bayesian Gaussian beta process. (a) Previous depth hypotheses before propagation. (b) Depth hypotheses after propagation and collision handling. Holes appear due to scale changes (move forward). (c) We fill holes using neighboring depth hypotheses. (d) Current depth estimation (Sect.~\ref{subsec:depth_estiimation}) with (e) corresponding captured image. We update propagated depth hypotheses via Bayesian inference: (f) Updated depth with (g) corresponding covariance and (h) inlier probability expectation. Colors in depth vary according to the distance from the environment surface to the camera. For covariance, brighter intensities indicate larger covariances, while for inlier probability expectation, brighter intensities indicate higher expectation.}
	\label{fig:gaussian_beta}
\end{figure*}
\subsubsection{Preliminaries}
\label{subsubsec:preliminaries}
We observe that there are a few outlier depth estimates obtained in the previous step due to occlusion, lack of texture, violation of photometric consistency, etc. Different from \cite{vi_mean} where outliers are not taken into account, we explicitly deal with outlier depth estimates. We assume that outlier depth estimates are uniformly distributed among the depth sample set $\Phi(L)$. We thus model the distribution of a depth estimate $d_\mathbf{u}^t$ ($d_\mathbf{u}^*$ at time instant $t$) as a Gaussian $+$ uniform mixture model distribution \cite{alpha_beta, REMODE}: a good depth estimate is normally distributed around a correct depth $z_\mathbf{u}$ with probability $\pi_\mathbf{u}$, while an outlier estimate is uniformly distributed within an interval $[z_{l}, z_{r}]$ with probability $1-\pi_\mathbf{u}$. The depth estimate probability density function of cases (a) and (b) in Sect.~\ref{subsubsec:D} is defined as

{\small
\begin{flalign}
p(d^t_\mathbf{u}| \pi_\mathbf{u}, z_\mathbf{u}) = \begin{cases}
\pi_\mathbf{u} \mathcal{N}(d^t_\mathbf{u} | z_\mathbf{u}, r_\mathbf{u}^2 ) +(1-\pi_\mathbf{u}) \mathcal{U}(d^t_\mathbf{u} | z_{l}, z_{r}), &\mbox{(a)}\\ 
(1-\pi_\mathbf{u}) \mathcal{U}(d^t_\mathbf{u} | z_{l}, z_{r}), &\mbox{(b)}
\end{cases} \nonumber
\end{flalign}
}where $\mathcal{N}(d^t_\mathbf{u} | z_\mathbf{u}, r_\mathbf{u}^2 ) $ is a Gaussian distribution with mean $z_\mathbf{u}$ and covariance $ r_\mathbf{u}^2$, and $\mathcal{U}(d^t_\mathbf{u} | z_{l}, z_{r})$ is a uniform distribution with $z_{l}$ and $z_{r}$ corresponding to the depth range of interest. The posterior of $z_\mathbf{u}$, $\pi_\mathbf{u}$ given $d_\mathbf{u}^t$ ($t \in [0 \ 1 \ ... \ n]$) is
\begin{flalign}
p( \pi_\mathbf{u}, z_\mathbf{u} | d_\mathbf{u}^{n}, ..., d_\mathbf{u}^0) &\propto  p( d_\mathbf{u}^{n}, ..., d_\mathbf{u}^0| \pi_\mathbf{u}, z_\mathbf{u}) p( \pi_\mathbf{u}, z_\mathbf{u}) \nonumber \\
&\propto p( d_\mathbf{u}^{n}, ..., d_\mathbf{u}^0| \pi_\mathbf{u}, z_\mathbf{u})  \nonumber \\
& = p( d_\mathbf{u}^{n}| \pi_\mathbf{u}, z_\mathbf{u}) p( d_\mathbf{u}^{n-1}, ..., d_\mathbf{u}^0| \pi_\mathbf{u}, z_\mathbf{u}) \nonumber \\ 
&\propto   p( d_\mathbf{u}^{n} | \pi_\mathbf{u}, z_\mathbf{u}) p( \pi_\mathbf{u}, z_\mathbf{u} | d_\mathbf{u}^{n-1}, ..., d_\mathbf{u}^0)
\end{flalign}
Similar to \cite{alpha_beta} and \cite{REMODE}, we approximate $p( \pi_\mathbf{u}, z_\mathbf{u} | d_\mathbf{u}^{n}, ..., d_\mathbf{u}^1, d_\mathbf{u}^0)$ using the product of a Gaussian distribution and a beta distribution for the sake of inference:
\begin{flalign}
\label{equ:q_fun}
& q(\pi_\mathbf{u}, z_\mathbf{u} | a_\mathbf{u}, b_\mathbf{u},  \mu_\mathbf{u}, \sigma_\mathbf{u} ) = \mathcal{N}(z_\mathbf{u} | \mu_\mathbf{u}, \sigma_\mathbf{u}^2 )  \mathcal{B}(\pi_\mathbf{u} | a_\mathbf{u}, b_\mathbf{u} ) 
\end{flalign}
where $\mu_\mathbf{u} $ and $\sigma_\mathbf{u}^2$ are the mean and variance of the depth estimate, while $ a_\mathbf{u}$ and $b_\mathbf{u}$ are probabilistic counters of how many inlier and outlier measurements have occurred during the lifetime of the depth estimate. This leads to:
\begin{flalign}
\label{equ:q_approximation}
& q(\pi_\mathbf{u}, z_\mathbf{u} | a_\mathbf{u}^{n}, b_\mathbf{u}^{n},  \mu_\mathbf{u}^{n}, {\sigma}^{n}_\mathbf{u} )   \approx  p( \pi_\mathbf{u}, z_\mathbf{u} | d_\mathbf{u}^{n}, ..., d_\mathbf{u}^1, d_\mathbf{u}^0) \nonumber \\
& \approx  p( d_\mathbf{u}^{n} | \pi_\mathbf{u}, z_\mathbf{u}) p(\pi_\mathbf{u}, z_\mathbf{u} | a^{n-1}_\mathbf{u}, b^{n-1}_\mathbf{u},  \mu^{n-1}_\mathbf{u}, \sigma^{n-1}_\mathbf{u} ) . \nonumber \\
& \approx  p( d_\mathbf{u}^{n} | \pi_\mathbf{u}, z_\mathbf{u}) q(\pi_\mathbf{u}, z_\mathbf{u} | a^{n-1}_\mathbf{u}, b^{n-1}_\mathbf{u},  \mu^{n-1}_\mathbf{u}, \sigma^{n-1}_\mathbf{u} ) .
\end{flalign}
We refer readers to \cite{alpha_beta} for details of the posterior update for case (a). The posterior update for case (b), however, is novel and not covered in \cite{alpha_beta} or \cite{REMODE}. We present the mathematic derivation in the following. Recall that the definition of beta function is:
\begin{flalign}
\label{equ:beta}
	 \mathcal{B}(\pi_\mathbf{u} | a_\mathbf{u}, b_\mathbf{u} ) = \frac{\Gamma (a_\mathbf{u} + b_\mathbf{u})}{\Gamma (a_\mathbf{u}) \Gamma (b_\mathbf{u}) } \pi_\mathbf{u}^{a_\mathbf{u}-1} (1-\pi_\mathbf{u})^{b_\mathbf{u}-1}
\end{flalign}
where $\Gamma(\cdot)$ is the gamma function. We increase $b_\mathbf{u}$ by 1, which leads to:
\begin{flalign}
\label{equ:beta}
\mathcal{B}(\pi_\mathbf{u} | a_\mathbf{u}, b_\mathbf{u} + 1 ) &= \frac{\Gamma (a_\mathbf{u} + b_\mathbf{u} +1 )}{\Gamma (a_\mathbf{u}) \Gamma (b_\mathbf{u} + 1) } \pi_\mathbf{u}^{a_\mathbf{u}-1} (1-\pi_\mathbf{u})^{b_\mathbf{u}} \nonumber \\
&= \frac{(a_\mathbf{u} + b_\mathbf{u}) \Gamma (a_\mathbf{u} + b_\mathbf{u} )}{b_\mathbf{u} \Gamma (a_\mathbf{u}) \Gamma (b_\mathbf{u} ) } \pi_\mathbf{u}^{a_\mathbf{u}-1} (1-\pi_\mathbf{u})^{b_\mathbf{u}} \nonumber \\
&= \frac{a_\mathbf{u} + b_\mathbf{u}}{b_\mathbf{u}} (1-\pi_\mathbf{u})  \mathcal{B}(\pi_\mathbf{u} | a_\mathbf{u}, b_\mathbf{u} ) 
\end{flalign}
where $\Gamma(a+1) = a \Gamma(a) $ is the property of gamma function. Substituting \eqref{equ:q_fun} and \eqref{equ:beta} into \eqref{equ:q_approximation}, we have
\begin{flalign}
&\mathcal{N}(z^n_\mathbf{u} | \mu^n_\mathbf{u}, {\sigma_\mathbf{u}^n}^2 )  \mathcal{B}(\pi_\mathbf{u} | a^n_\mathbf{u}, b^n_\mathbf{u} ) \nonumber \\ 
\approx \ &  (1-\pi_\mathbf{u}) \mathcal{U}(d^t_\mathbf{u} | z_{l}, z_{r}) \mathcal{N}(z_\mathbf{u} | \mu^{n-1}_\mathbf{u}, {\sigma_\mathbf{u}^{n-1}}^2 )  \mathcal{B}(\pi_\mathbf{u} | a^{n-1}_\mathbf{u}, b^{n-1}_\mathbf{u} ) \nonumber \\
\propto \ & \frac{b^{n-1}_\mathbf{u}}{a^{n-1}_\mathbf{u} + b^{n-1}_\mathbf{u}} \mathcal{N}(z_\mathbf{u} | \mu^{n-1}_\mathbf{u}, {\sigma_\mathbf{u}^{n-1}}^2 )  \mathcal{B}(\pi_\mathbf{u} | a^{n-1}_\mathbf{u}, b^{n-1}_\mathbf{u} + 1 ) \nonumber \\
\propto \ & \mathcal{N}(z_\mathbf{u} | \mu^{n-1}_\mathbf{u}, {\sigma_\mathbf{u}^{n-1}}^2 )  \mathcal{B}(\pi_\mathbf{u} | a^{n-1}_\mathbf{u}, b^{n-1}_\mathbf{u} + 1 )
\end{flalign}
which yields
\begin{flalign}
\mu^n_\mathbf{u} = \mu^{n-1}_\mathbf{u}, \ \sigma_\mathbf{u}^n = \sigma_\mathbf{u}^{n-1}, \  a^{n}_\mathbf{u} = a^{n-1}_\mathbf{u}, \ \ b^{n}_\mathbf{u} = b^{n-1}_\mathbf{u} + 1.
\end{flalign}


\subsubsection{Recursive Estimation}
In contrast to \cite{alpha_beta} and \cite{REMODE}, where the temporal and spatial correlations of consecutive depth estimates are ignored, we make use of these correlations. Each depth hypothesis consists of three variables: mean, covariance and inlier probability expectation. We update depth hypotheses in a recursive way.
An illustration of the proposed Bayesian Gaussian beta process is shown in Fig.~\ref{fig:gaussian_beta}. Details are as follows:

\noindent\textbf{Initialization}:
For the first depth estimate, we initialize the depth hypothesis of pixel $\mathbf{u}$: $a_\mathbf{u}^{0} =  b_\mathbf{u}^{0} = 10$,  $\mu_\mathbf{u}^{0} = d_\mathbf{u}^{0}$, and ${{\sigma}^{0}_\mathbf{u}}^2 = (\frac{\partial \ \frac{1}{disp \cdot c_d}}{\partial \ disp})^2 \sigma_{disp}^2 $ with $disp = \frac{1}{d_\mathbf{u}^{0} \cdot c_d}$ as well as $\sigma_{disp}^2 = 1$. This step is only performed once at the beginning of the Bayesian Gaussian beta process.

\noindent\textbf{Propagation}:
We propagate the depth hypothesis from the previous reference frame $a^{n-1}_\mathbf{u}, b^{n-1}_\mathbf{u},  \mu^{n-1}_\mathbf{u}, \sigma^{n-1}_\mathbf{u} $ to the new reference frame $a_\mathbf{u'}^{n-1}, b_\mathbf{u'}^{n-1},  \mu_\mathbf{u'}^{n-1}, {\sigma}^{n-1}_\mathbf{u'}$. Assuming the rotation is small, we have 
\begin{flalign}
\mu^{n-1}_\mathbf{u'} &= \mu^{n-1}_\mathbf{u} - t_z, \ \ {\sigma^{n-1}_\mathbf{u'}}^2 = {\sigma^{n-1}_\mathbf{u} }^2 + \sigma_{t_z}^2 \\
a^{n-1}_\mathbf{u'} &= a^{n-1}_\mathbf{u}, \ \ \ \ \ \ \ \ \ \ \ b^{n-1}_\mathbf{u'} = b^{n-1}_\mathbf{u}
\end{flalign}
where $t_z$ is the translation perpendicular to the camera plane and $\sigma_{t_z}^2$ is the variance of $t_z$. For simplicity, we set $\sigma_{t_z}^2$ to be $0.05^2$ in this work. We do not propagate depth hypotheses whose inlier probability expectation $E[\mathcal{B}(\pi_\mathbf{u} | a^{n-1}_\mathbf{u}, b^{n-1}_\mathbf{u} )$] = $\frac{a^{n-1}_\mathbf{u}}{a^{n-1}_\mathbf{u}+b^{n-1}_\mathbf{u} } $ is less than 0.4 (i.e., it is unlikely to be an inlier depth estimate).

\noindent\textbf{Collision Handling}:
At all times, we allow at most one depth hypothesis per pixel. However, this is not the case for the scale change (i.e., move backward) as well as occlusion. If two or more depth hypothesis are propagated to the same pixel in the new keyframe, we save the depth hypotheses whose inlier probability expectation $E[\mathcal{B}(\pi_\mathbf{u} | a^{n-1}_\mathbf{u}, b^{n-1}_\mathbf{u} )$] = $\frac{a^{n-1}_\mathbf{u}}{a^{n-1}_\mathbf{u}+b^{n-1}_\mathbf{u} } $ is larger than 0.5  (i.e., not likely to be an outlier depth estimate) as well as whose mean $\mu^{n-1}_\mathbf{u} $ is the smallest (for occlusion handling).

\noindent\textbf{Hole Filling}:
Due to scale changes (i.e., move forward), holes may appear after propagation. We set each depth hypothesis in the holes to be the same as its nearest neighbors with distance less than $\tau_d$ pixels. Threshold $\tau_d$ balances the similarity between neighboring depth hypotheses against the variation: a large $\tau_d$ helps to fill more holes at the cost of less accurate depth hypotheses, while a small $\tau_d$ leads to more holes but accurate depth hypotheses. We empirically set $\tau_d$ to be 2 in this work.

\noindent\textbf{Update}:
If the depth hypothesis of pixel $\mathbf{u}$ is null after propagation and hole filling, we initialize it as $a_\mathbf{u}^{n} =  b_\mathbf{u}^{n} = 10$,  $\mu_\mathbf{u}^{n} = d_\mathbf{u}^{n}$, and ${{\sigma}^{n}_\mathbf{u}}^2 = (\frac{\partial \ \frac{1}{disp \cdot c_d}}{\partial \ disp})^2 \sigma_{disp}^2 $ with $disp = \frac{1}{d_\mathbf{u}^{n} \cdot c_d}$ as well as $\sigma_{disp}^2 = 1$. Otherwise, we update its posterior distribution of $q(\pi_\mathbf{u}, z_\mathbf{u} | a_\mathbf{u}^{n}, b_\mathbf{u}^{n},  \mu_\mathbf{u}^{n}, {\sigma}^{n}_\mathbf{u} ) $ according to update formulations mentioned in Sect.~\ref{subsubsec:preliminaries}. 

\noindent\textbf{Output}:
For each pixel $\mathbf{u}$, we output its mean $\mu^n_\mathbf{u}$, variance ${\sigma_\mathbf{u}^n}^2$ and inlier probability expectation $E[\mathcal{B}(\pi_\mathbf{u} | a^{n}_\mathbf{u}, b^{n}_\mathbf{u} )$] = $\frac{a^{n}_\mathbf{u}}{a^{n}_\mathbf{u}+b^{n}_\mathbf{u} } $ if $E[\mathcal{B}(\pi_\mathbf{u} | a^{n}_\mathbf{u}, b^{n}_\mathbf{u} )] > 0.6$. These outputs are needed for the uncertainty-aware depth integration to be discussed in Sect.~\ref{subsec:depth_integration}.


\subsection{Uncertainty-aware Depth Integration}
\label{subsec:depth_integration}

To build compact and dense 3D models, we adopt the idea of volumetric fusion \cite{kinectfusion,Niebner2013,Whelan16ijrr} to integrate all depth estimates obtained in the previous subsection. In contrast to \cite{kinectfusion,Niebner2013,Whelan16ijrr}, where dense reconstructions are based on light-emitting depth cameras that provide high-quality measurements, depth estimation from a moving camera contains noticeable outliers. This motivates us to explicitly model the inlier probability of each depth estimate in the previous subsection and take outliers into account in the depth integration step. 

We represent the world as a 3D array of cubic voxels. Each voxel is associated with a signed distance function (SDF) $\phi( \mathbf{x} ): \mathbb{R}^3 \rightarrow \mathbb{R} $ and a weight $w(\mathbf{x}): \mathbb{R}^3 \rightarrow \mathbb{R} $. SDF
$\phi( \mathbf{x})$ denotes the signed distance between $\mathbf{x}$ and the nearest object surface, and it is positive if it is outside an object and negative otherwise.
It can be easily seen that surfaces of objects are zero crossings of signed distance functions (i.e., $\phi( \mathbf{x} ) = 0$). 
$w(\mathbf{x})$ represents the confidence of the sigined distance function. As shown in \cite{BCML96}, averaging distance measurements with respective variances over time results in minimizing the weighted sum of the square distances to all ray endpoints for the zero isosurface of the SDF. 

Since the major part of the 3D world is usually empty, we use a hash table to index voxels and only store SDFs, as well as their weights, that are near object surfaces \cite{Klingensmith2015,Niebner2013}. These SDFs are called truncated signed distance functions (TSDFs):
\begin{flalign}
\phi_r( \mathbf{x} )&=  \begin{cases}
\phi( \mathbf{x} ), & \text{if  } \ ||\phi( \mathbf{x} )|| \leq  r \\
\text{undefined}, & \text{otherwise}
\end{cases},
\end{flalign}
where $r$ is the truncated distance threshold. For a given depth measurement $d$ with corresponding ray vector direction $\mathbf{f}$, we classify segements of a ray into three regions\cite{Klingensmith2015}:
\begin{flalign}
u \cdot \mathbf{f} \in \begin{cases}
\text{hit region}, &  \text{if} \ \ || u  - d || \leq  r \\
\text{space carving region},  & \text{if} \ \ u \leq d - r \\
\text{undefined}, & \text{otherwise}
\end{cases}.
\end{flalign}

\subsubsection{Uncertainty-aware TSDF Update}
Voxels within the hit region are updated as:
\begin{flalign}
\phi_r( \mathbf{x} )' &= \frac{\phi_r( \mathbf{x} ) \cdot w(\mathbf{x}) + \delta d \cdot \alpha( \delta d ) }{w(\mathbf{x}) + \alpha(\delta d) } \\
w(\mathbf{x})' &= w(\mathbf{x}) + \alpha(\delta d) 
\end{flalign}
where $\delta d = \mathbf{x}-d \cdot \mathbf{f}$, and $\alpha (\delta d) $ is the corresponding variance obtained in Sect.~\ref{subsec:probabilistic_filtering}. While $\alpha (\delta d) $ in \cite{kinectfusion,Niebner2013,Whelan16ijrr} is a constant, we set it to be the variance obtained from hypothesis filtering to take the uncertainty of motion stereo into account. The initial condition of a TSDF is $\phi_r( \mathbf{x} ) = constant $ and $w(\mathbf{w}) = 0$.

\subsubsection{Uncertainty-aware Ray Tracing}
Voxels within the space carving region are chiseled away.
This operation can be viewed as removing potential depth outliers by visibility constraints (i.e., segments between two endpoints of a ray are empty).
Free-space carving makes sense for the reason that we care more about which part of the scene does not contain surfaces (for motion planning) than what is inside objects.
However, free-space carving with outlier depth measurements is harmful to the built model, as part of it may be wrongly chiseled away. Therefore, we only do ray tracing if the expectation of inlier probability obtained in Sect.~\ref{subsec:probabilistic_filtering} is large than 0.8.


While voxels are sufficient for motion planning \cite{CHOMP,HelenOleynikova16}, colors and textures are more suitable for visualization and debugging. We include an optional step, marching cubes \cite{marching_cubes}, to extract polygonal meshes of an isosurface from a three-dimensional discrete scalar field.

\begin{table*}[!h]
	\label{tab:time}
	\caption[Comparision of the avarage computation time on the TUM RGB-D SLAM \& IC-NUIM Dataset.]{Comparision of the avarage computation time on the TUM RGB-D SLAM \& IC-NUIM Dataset.}
	\centering
	\scalebox{0.85}{
		\begin{tabular}{|c| c| c| c | c | c | c|}
			\hline 
			{\bf Methods}  & {\bf Ours-T} & {\bf Ours-T+S} & {\bf Ours-T+S+D} & {\bf Ours-T+S+D+H} & {\bf REMODE} \cite{REMODE}& {\bf VI-MEAN} \cite{vi_mean}\\ \hline
			Average computation time (ms) & 33.71 & 41.20 &  42.10 & 51.02 & {31.31} & 80.02\\
			\hline
		\end{tabular}
	}
\end{table*}

\begin{table*}[!h]
	\label{tab:valid_mapping_density}
	\caption[Comparision of the average mapping density on the TUM RGB-D SLAM Dataset \& ICL-NUIM Dataset.]{Comparision of the average mapping density on the TUM RGB-D SLAM Dataset \& ICL-NUIM Dataset..}	
	\centering
	\scalebox{0.85}{
		\begin{tabular}{|c|c | c| c| c| c | c | c|}
			\hline 
			{\bf Dataset Name} &  {\bf Sequence Name}  & {\bf Ours-T} & {\bf Ours-T+S} & {\bf Ours-T+S+D} & {\bf Ours-T+S+D+H} & {\bf REMODE} \cite{REMODE}& {\bf VI-MEAN} \cite{vi_mean}\\ \hline
			\multirow{6}{*}{TUM RGB-D SLAM} & freiburg2\_desk & 60.57 &  64.14  & 62.34  & 46.52 & 32.63 & 85.44 \\
			& freiburg3\_nostructure\_texture\_far  & 65.42 & 76.68 & 74.62 & 71.40 & 44.62 & 69.13  \\
			& freiburg3\_sitting\_halfsphere & 67.59 & 69.80 & 66.17 & 61.89 & 22.29 & 56.60 \\  
			& freiburg3\_structure\_texture\_far & 80.26 & 87.05 & 86.72 & 80.88 & 34.16 & 76.26  \\
			& freiburg3\_structure\_notexture\_far & 76.68 & 85.49 & 84.05 & 82.28& 43.85 & 80.30 \\
			& freiburg3\_sitting\_xyz& 67.37 & 70.59 & 67.05&65.16& 22.61 & 51.32\\
			\hline
			\multirow{8}{*}{ICL-NUIM}  
     		& living room of kt0 & 87.26 & 91.82 & 90.01 & 88.93 & 68.80 & 96.05 \\  
			& living room of kt1 & 91.40 & 93.51 & 93.17 & 90.52 & 67.10 & 97.00 \\
			& living room of kt2 & 88.58 & 90.00 & 89.46 & 88.49 & 67.13 & 94.45 \\
			& living room of kt3 & 91.56 & 93.61 & 92.33 & 91.90 & 62.32 & 86.90 \\
			& office room of kt0 & 89.67 & 93.34 & 91.37 & 87.45 & 32.08 & 90.57 \\  
			& office room of kt1  & 90.09 & 95.55 & 93.90 & 88.90 & 25.24 & 95.66 \\  
			& office room of kt2 & 89.53 & 92.08 & 91.66 & 87.85 & 39.35 & 94.15 \\  
			& office room of kt3 & 91.35 & 96.12 & 93.37 & 89.22 & 27.87 & 93.50 \\  
			\hline
		\end{tabular}
	}
\end{table*}

\section{Experiments}
\label{sec:experiments}
The whole system is implemented in C++, with ROS as the interfacing robotics middleware. All testings are carried out in a commodity Lenovo laptop Y50 with an i7-4720HQ CPU and a mobile GTX-960M GPU.
The depth estimation module is run on the GPU while the hypothesis filtering module and the uncertainty-aware depth integration module are implemented in the CPU. These modules are placed on different threads to utilize the multi-core CPU architecture. 

\subsection{TUM RGB-D SLAM Dataset \& ICL-NUIM Dataset}
\begin{table}[!h]
	\label{tab:table_1}
	\centering
	\caption[Statistics of selected sequences on the TUM RGB-D SLAM Dataset.]{ \scriptsize{Statistics of selected sequences on the TUM RGB-D SLAM Dataset.} }
	\scalebox{0.9}{
		\begin{tabular}{|c | c | c | c|}
			\hline 
			{\bf Sequence Name}  & {\bf Structure} & {\bf Texture} & {\bf Dynamic Objects} \\ \hline
			freiburg2\_desk & $\surd$ & $\surd$ &  $\times$\\
			freiburg3\_nostructure\_texture\_far  & $\times$ & $\surd$ & $\times$ \\
			freiburg3\_sitting\_halfsphere & $\surd$ & $\surd$ & $\surd$ \\  
			freiburg3\_structure\_texture\_far & $\surd$ & $\surd$ & $\times$  \\
			freiburg3\_structure\_notexture\_far & $\surd$ & $\times$ & $\times$\\
			freiburg3\_sitting\_xyz & $\surd$ & $\surd$ & $\surd$\\
			\hline
		\end{tabular}
	}
\end{table}
\begin{figure}[!h]
	\centering
	\subfigure[ \scriptsize{freiburg2\_desk} ]{\includegraphics[width=0.45\columnwidth]{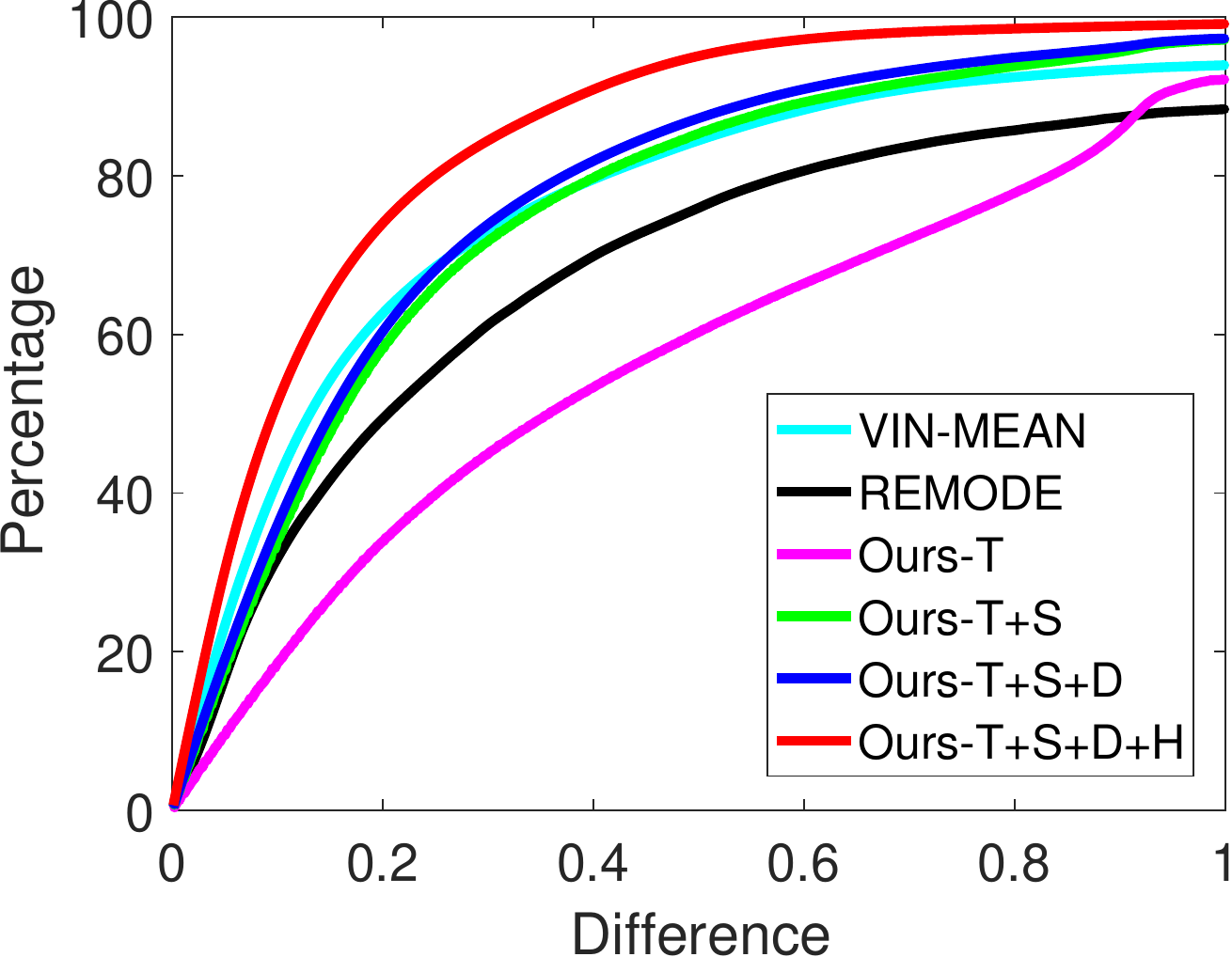}} 
	\subfigure[ \scriptsize{freiburg3\_nostructure\_texture\_far} ]{\includegraphics[width=0.45\columnwidth]{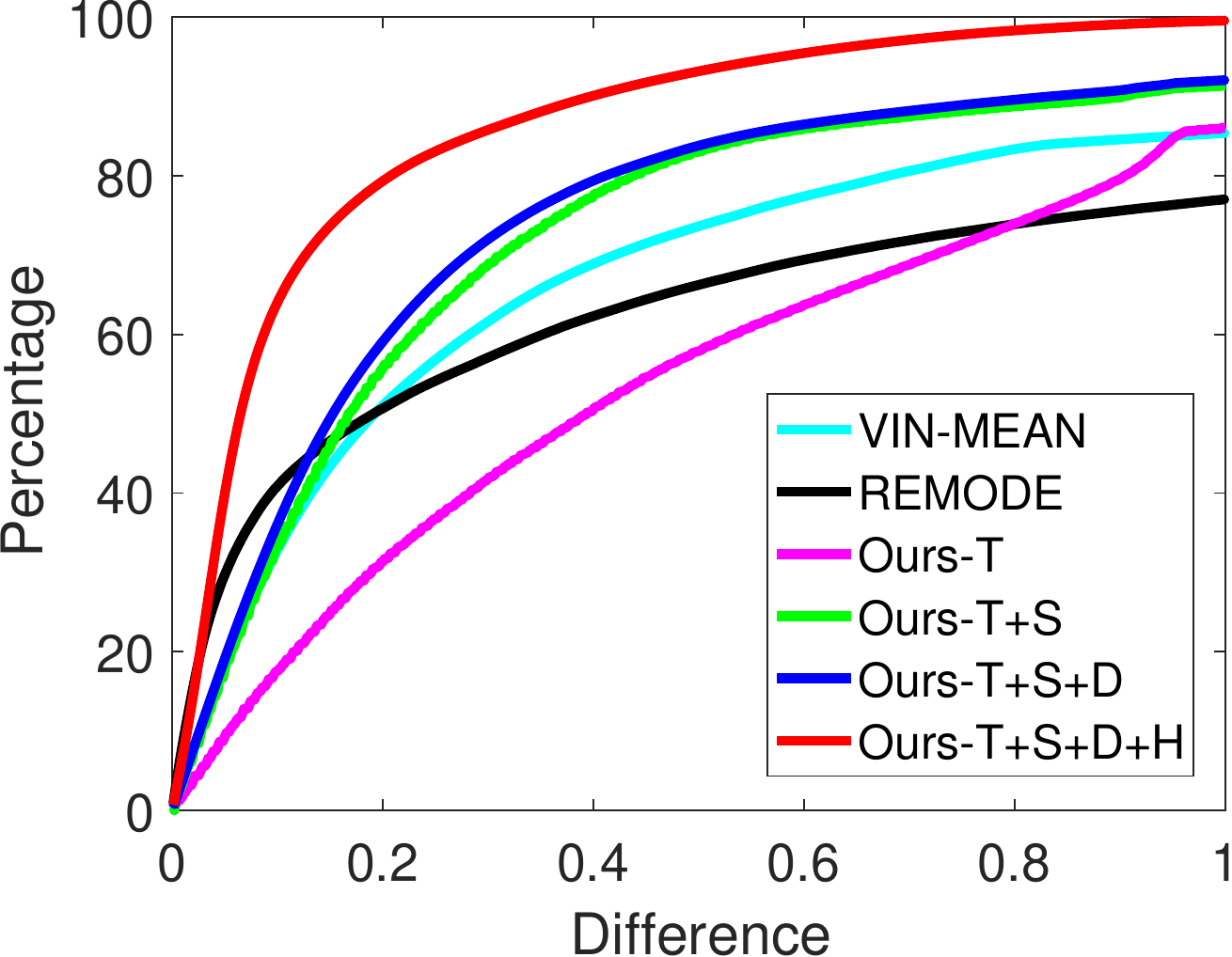}}
	\subfigure[ \scriptsize{ freiburg3\_sitting\_halfsphere} ]{\includegraphics[width=0.45\columnwidth]{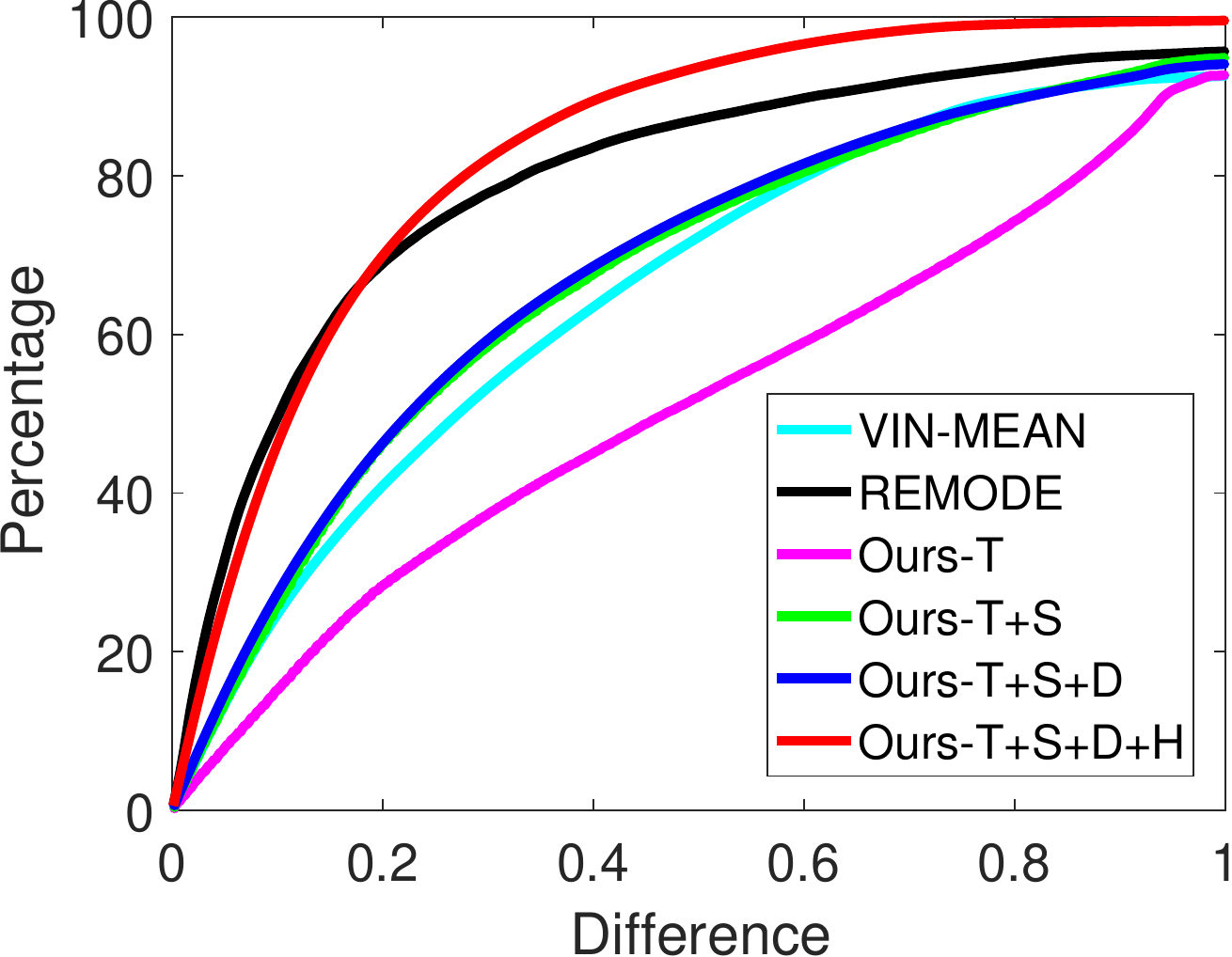}}
	\subfigure[ \scriptsize{freiburg3\_structure\_texture\_far} ]{\includegraphics[width=0.45\columnwidth]{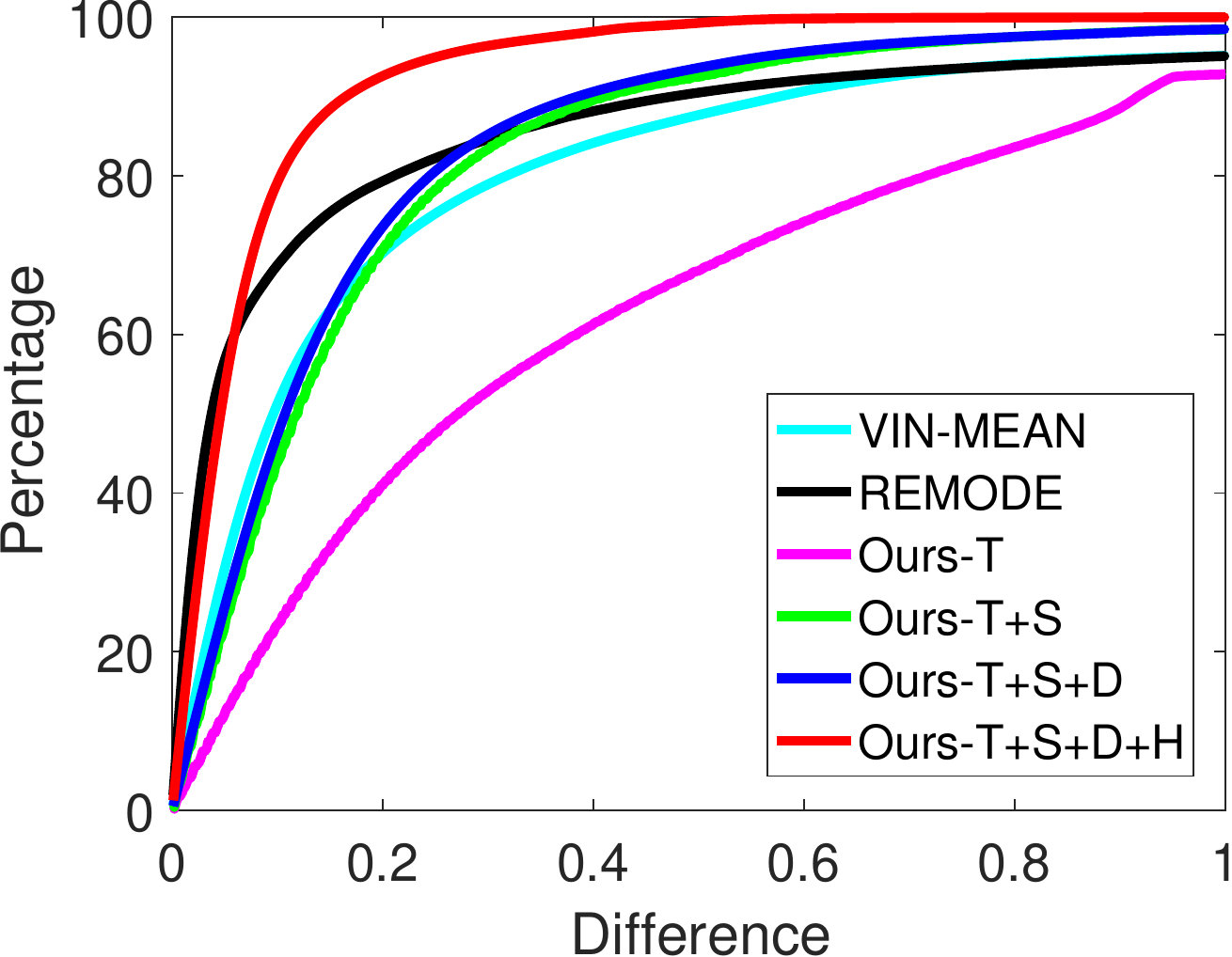}}
	\subfigure[ \scriptsize{freiburg3\_structure\_notexture\_far} ]{\includegraphics[width=0.45\columnwidth]{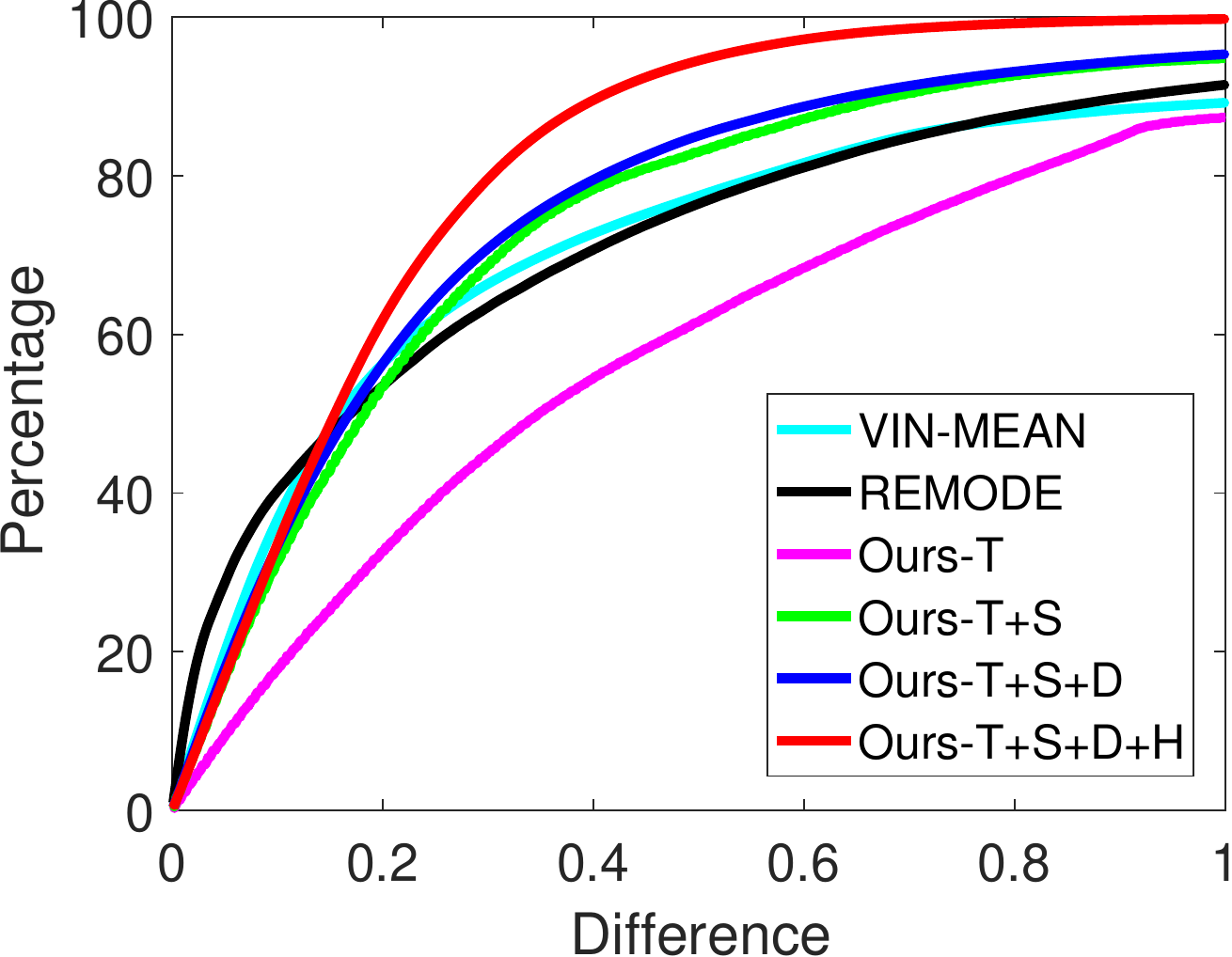}}
	\subfigure[ \scriptsize{freiburg3\_sitting\_xyz} ]{\includegraphics[width=0.45\columnwidth]{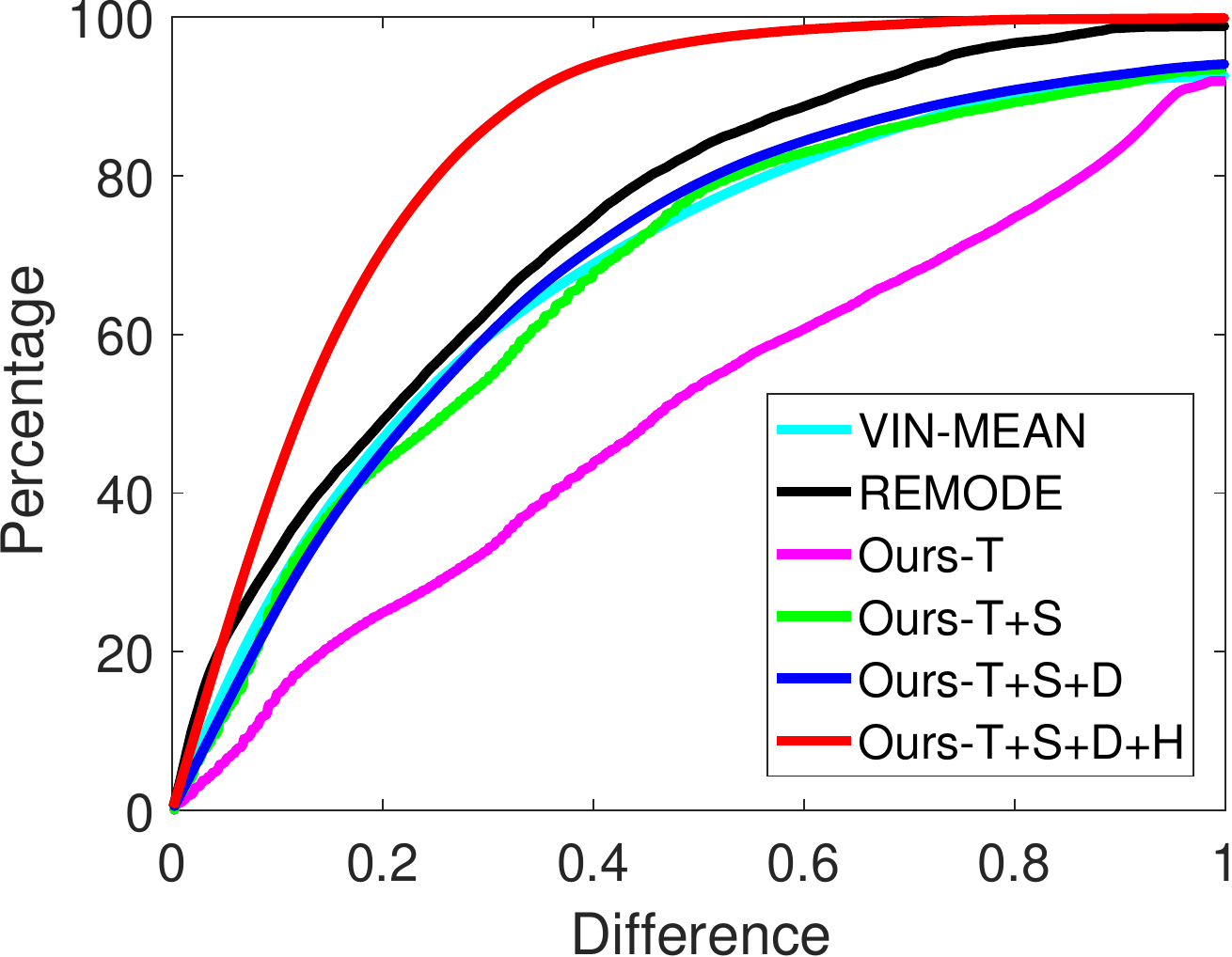}}
	\caption[Comparision of per-depth error percentage (\% w.r.t. m) in sequences of the TUM RGB-D SLAM dataset.]{Comparision of per-depth error percentage (\% w.r.t. m) in sequences of the TUM RGB-D SLAM dataset. We calculate the percentage (vertical axis) of depth difference, between the estimated depth values and the ground truth depth values, within the difference threshold $e_d$ (horizontal axis). Our approach (T+S+D+H) achieves higher mapping accuracy than state-of-the-art methods (REMODE \cite{REMODE} and VI-MEAN \cite{vi_mean}).}
	\label{fig:mapping_accuracy_tum.}
\end{figure}
We evaluate the mapping performance of our monocular depth estimation obtained after hypothesis filtering on the TUM RGB-D SLAM dataset \footnote{ \scriptsize{\url{https://vision.in.tum.de/data/datasets/rgbd-dataset}} } and the ICL-NUIM dataset \footnote{ \scriptsize{\url{https://www.doc.ic.ac.uk/~ahanda/VaFRIC/iclnuim.html}} }. We use ground truth poses from datasets as mapping pose inputs to ensure correct mapping metric for evaluation. Depths from Microsoft Kinect (TUM RGB-D SLAM dataset) or ray tracking (ICL-NUIM dataset) are used for mapping performance evaluation. Since the TUM RGB-D SLAM dataset is originally for odometry, we select some static sequences that are suitable for dense mapping. These selected sequences cover various environment conditions (Table III). We use an ablation study for analysis: T denotes temporal cost aggregation (Sect.~\ref{subsubsec:T}); S denotes spatial regulation (Sect.~\ref{subsubsec:S}); D denotes local region discussion \& depth refinement (Sect.~\ref{subsubsec:D}); H denotes hypothesis filtering (Sect.~\ref{subsec:probabilistic_filtering}). We also compare our approach with state-of-the-art methods: REMODE \cite{REMODE} and VI-MEAN \cite{vi_mean}. Three measurement metrics are used for comparison:
\begin{figure*}[!ht]
	\centering
	\subfigure[living room of kt0]{\includegraphics[width=0.45\columnwidth]{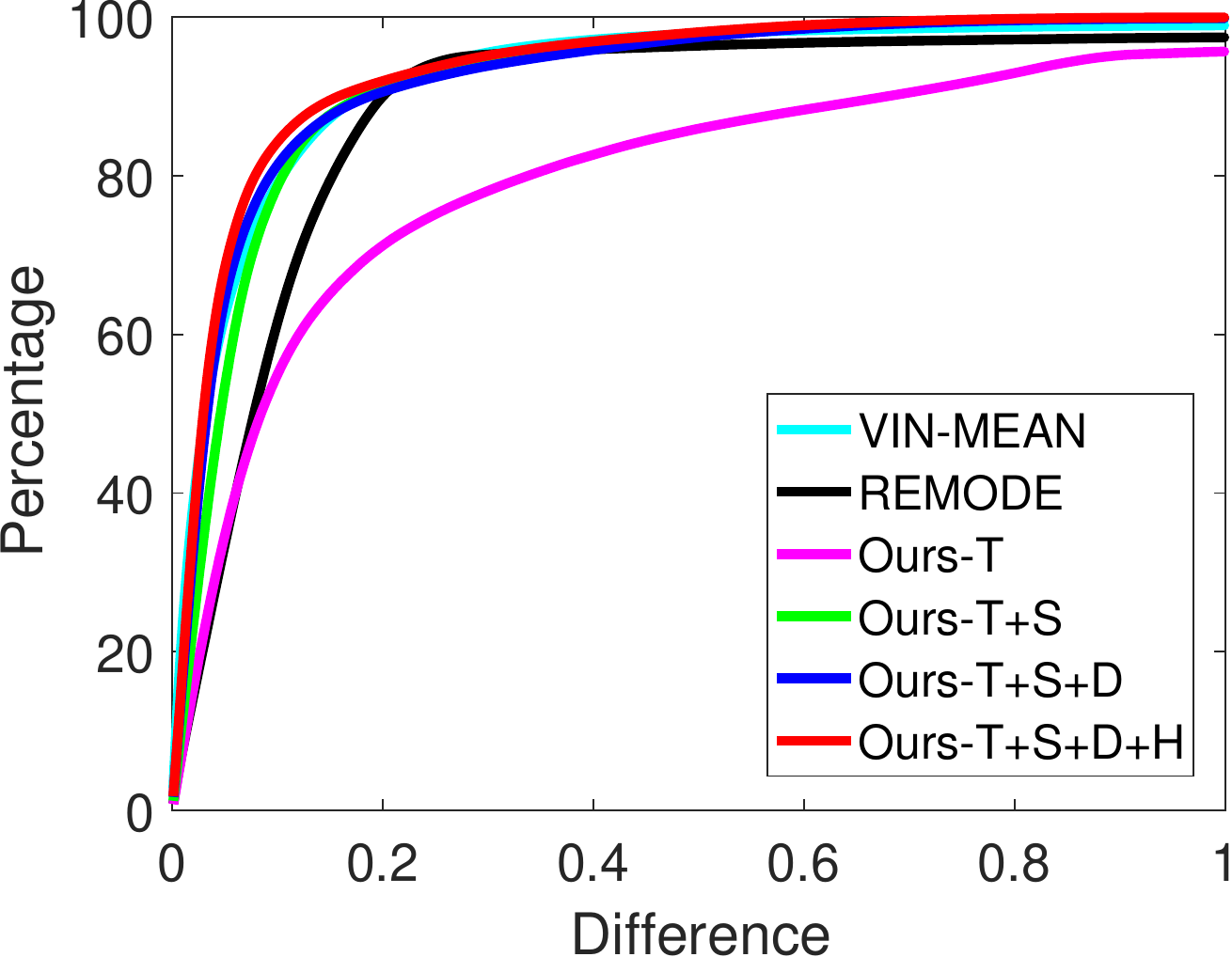}} \ \
	\subfigure[living room of kt1]{\includegraphics[width=0.45\columnwidth]{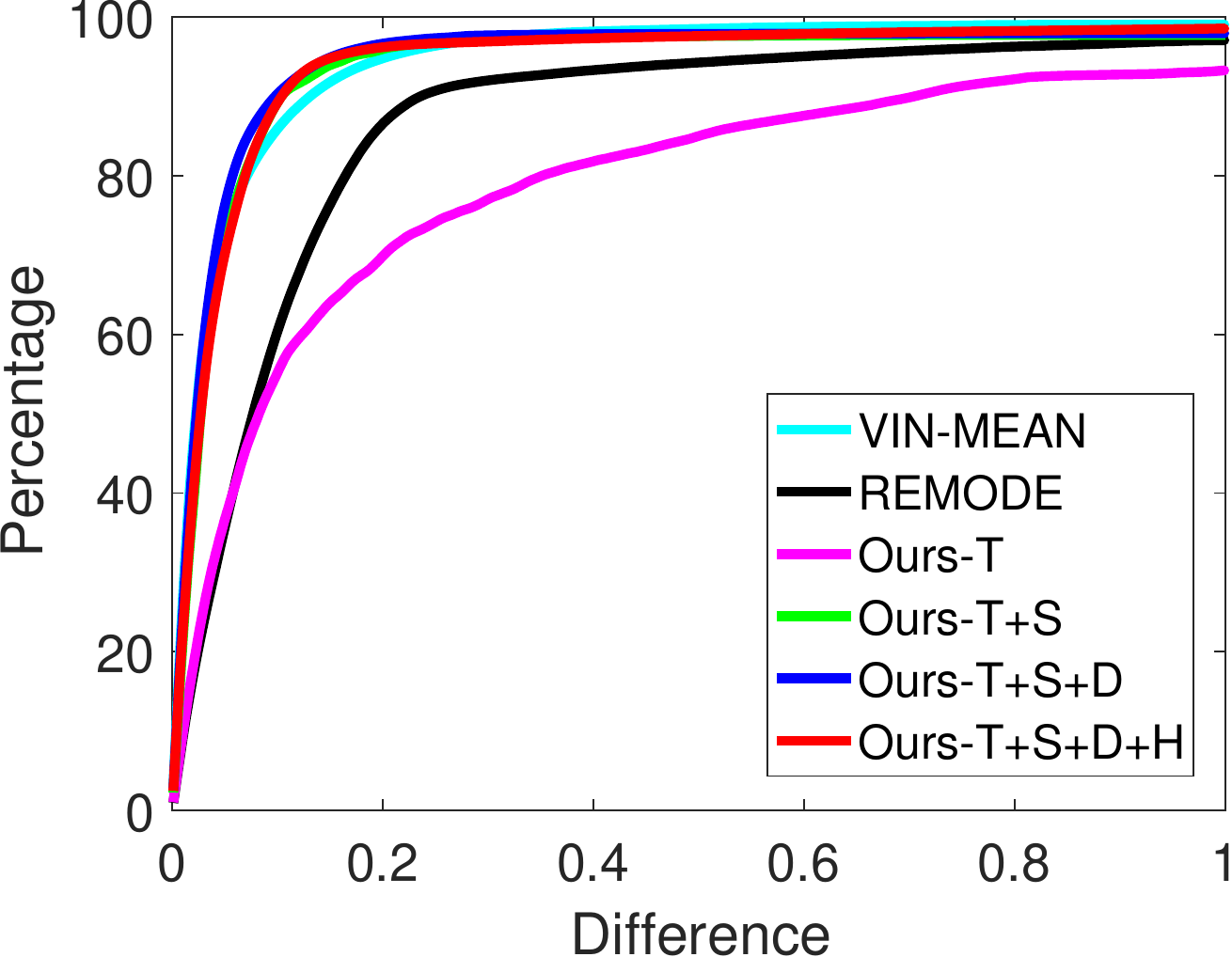}} \ \
	\subfigure[living room of kt2]{\includegraphics[width=0.45\columnwidth]{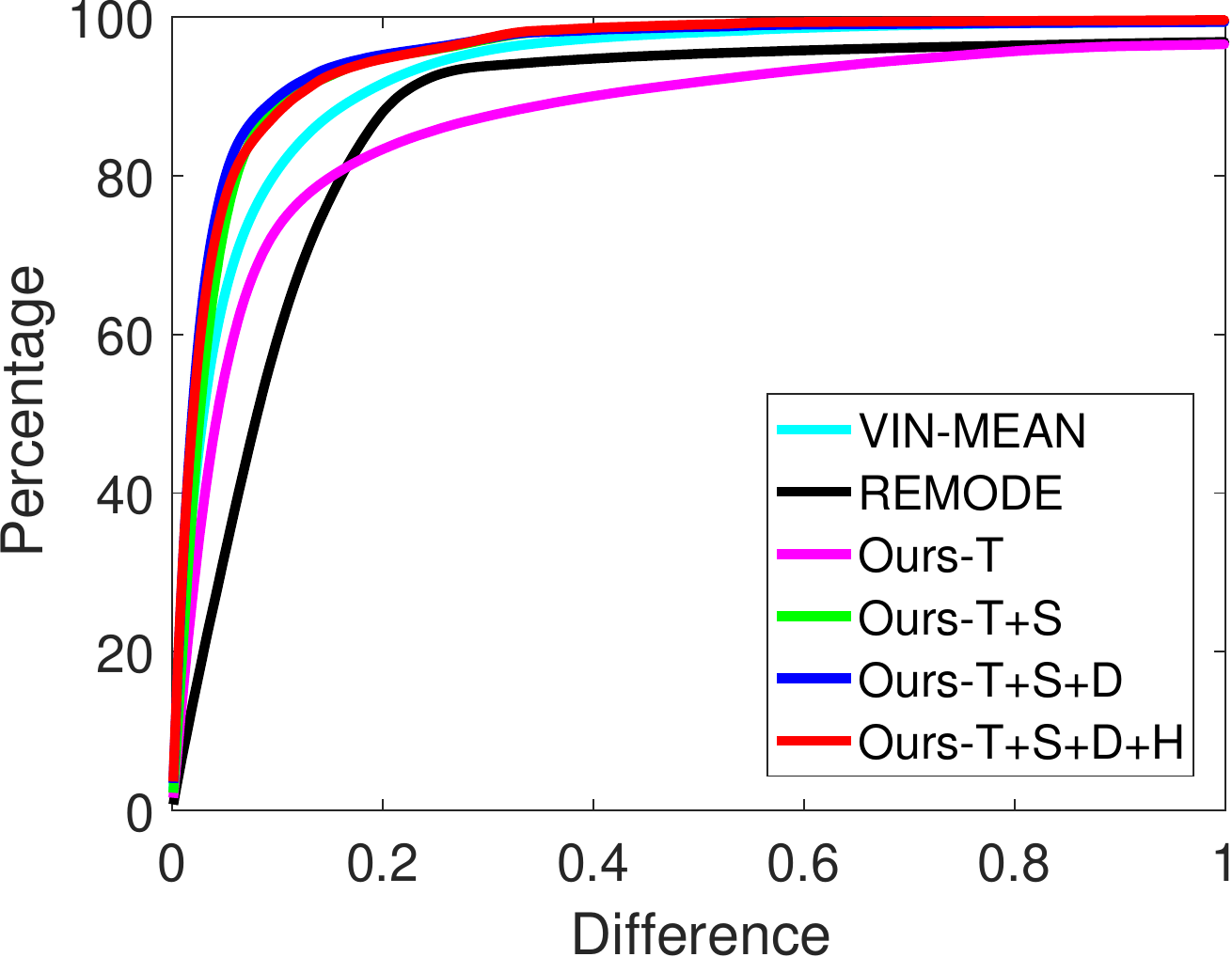}} \ \
	\subfigure[living room of kt3]{\includegraphics[width=0.45\columnwidth]{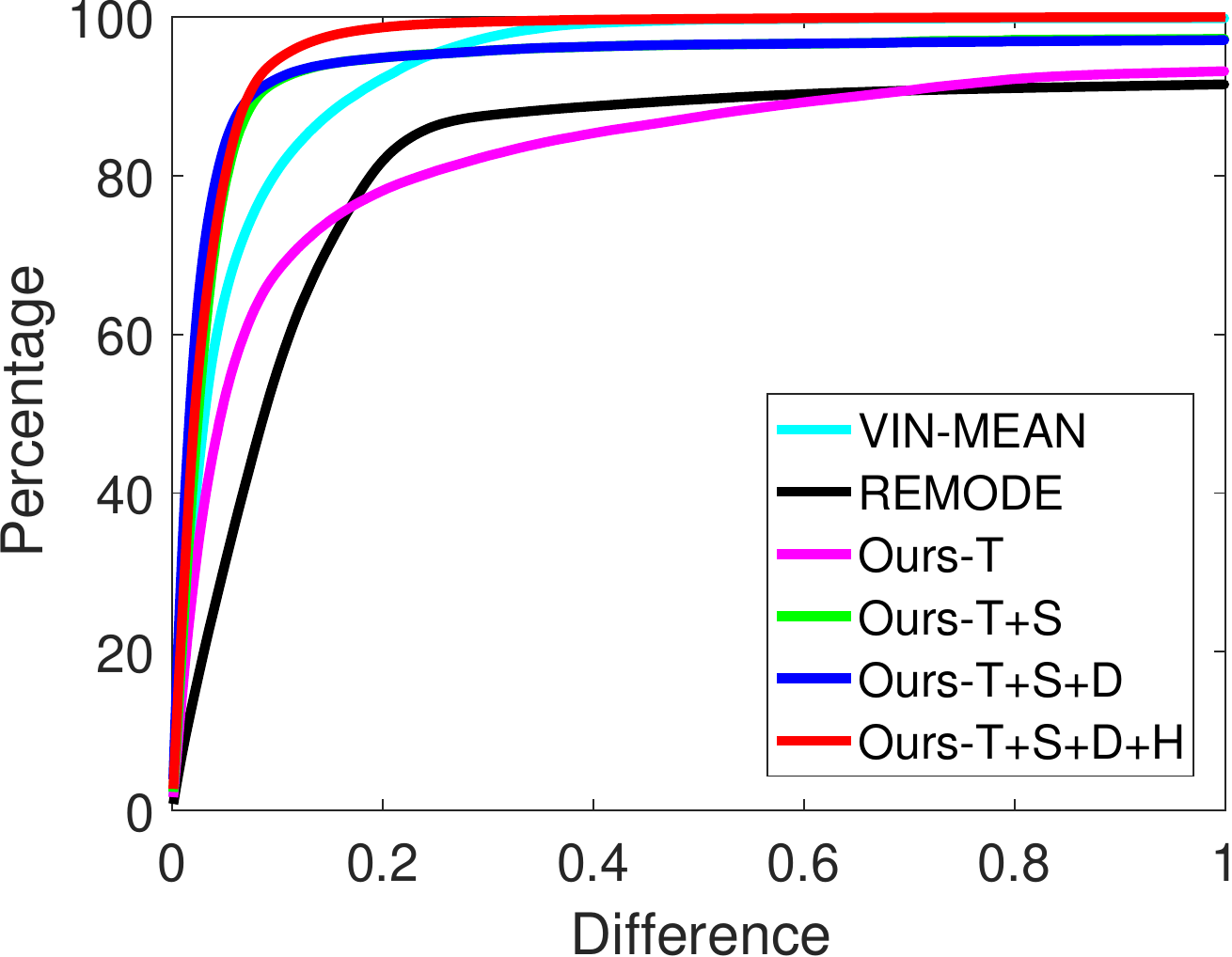}} 
	\subfigure[office room of kt0]{\includegraphics[width=0.45\columnwidth]{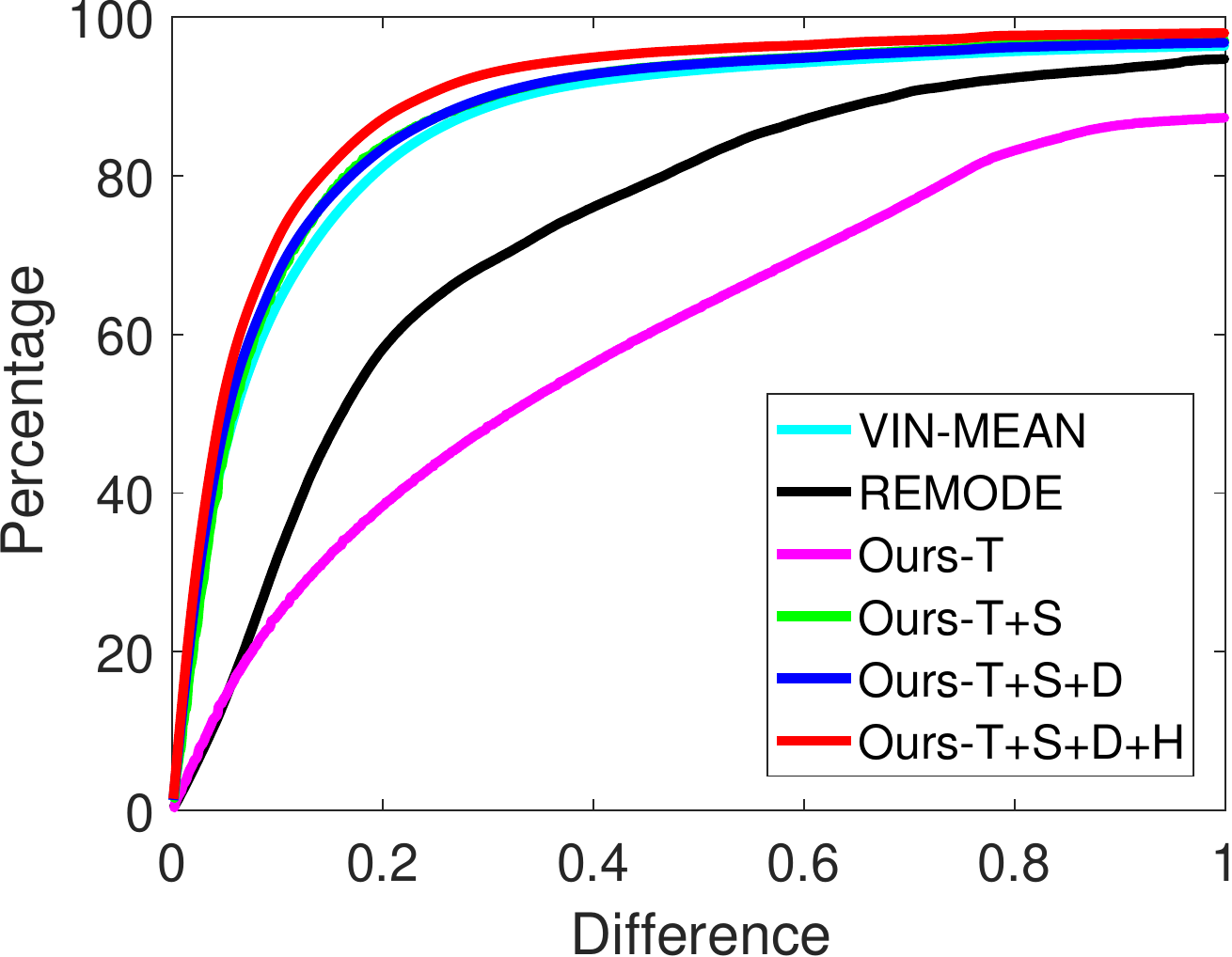}} \ \
	\subfigure[office room of kt1]{\includegraphics[width=0.45\columnwidth]{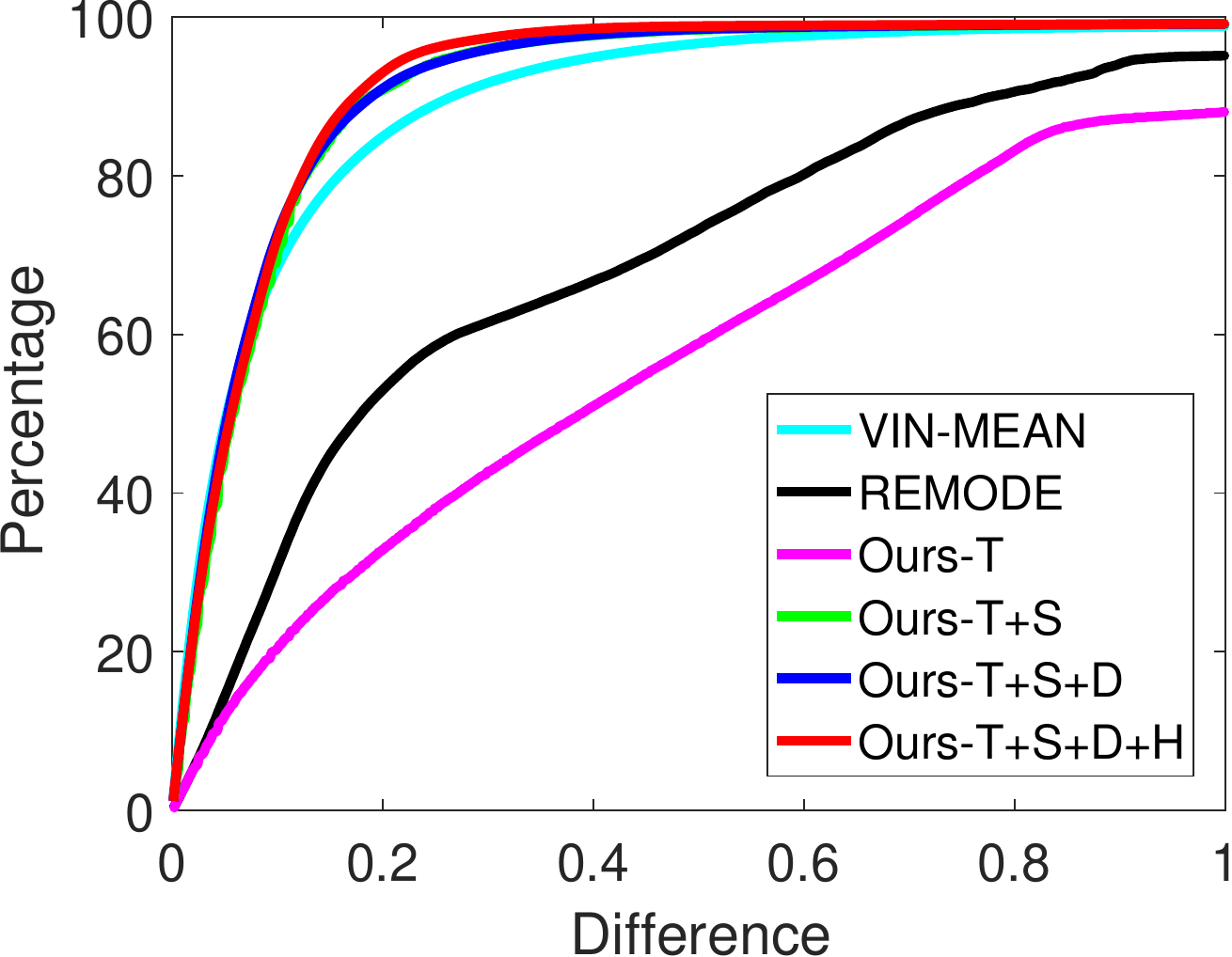}} \ \
	\subfigure[office room of kt2]{\includegraphics[width=0.45\columnwidth]{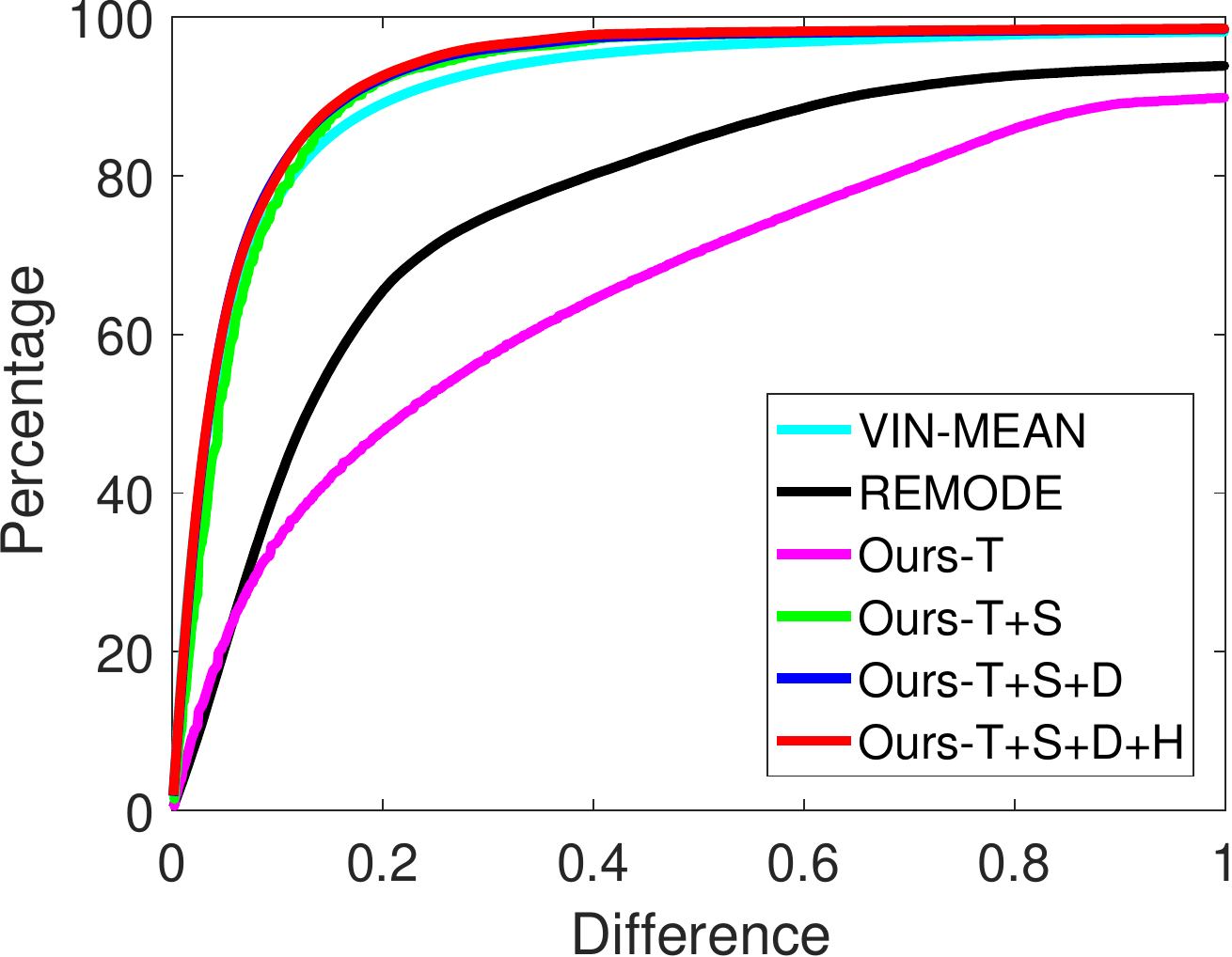}} \ \
	\subfigure[office room of kt3]{\includegraphics[width=0.45\columnwidth]{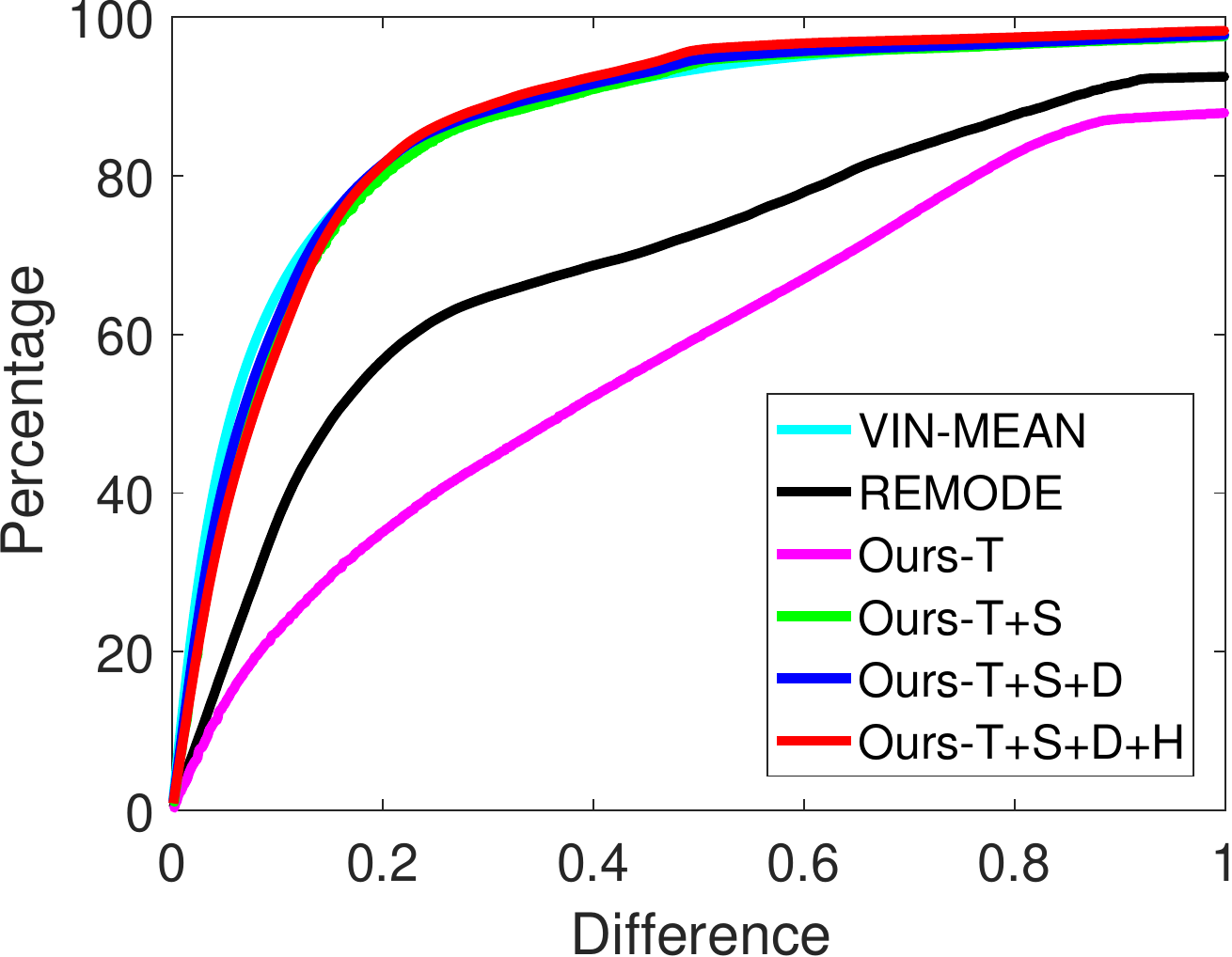}}
	\caption[Comparision of per-depth error percentage (\% w.r.t. m) in sequences of the ICL-NUIM dataset.]{Comparision of per-depth error percentage (\% w.r.t. m) in sequences of the ICL-NUIM dataset. We calculate the percentage (vertical axis) of depth difference, between the estimated depth values and the ground truth depth values, within the difference threshold $e_d$ (horizontal axis). Our approach (T+S+D+H) achieves higher mapping accuracy than state-of-the-art methods (REMODE \cite{REMODE} and VI-MEAN \cite{vi_mean}).}
	\label{fig:mapping_accuracy_icl.}
\end{figure*}

\begin{itemize}
	\item Average computation time (ms): the average computation time of each depth computation. It evaluates the online performance of mobile applications.
	\item Average mapping density (\%): the average density of depth estimates for each depth estimation. It plays a key role in the safety of mobile robots. A higher density helps better obstacle avoidance.
	\item Per-depth error percentage (\% w.r.t. m): the percentage of depth difference, between the estimated depth values and the ground truth depth values, within the difference threshold $e_d$. It evaluates mapping accuracy. We prefer higher percentage of small estimation errors.
\end{itemize}

Since the image resolution of sequences on both datasets are the same, the computation times of different approaches on different sequences are similar. We take the average of the computation times, and summarize them in Table I. 

\begin{figure*}[!ht]
	\centering
	\subfigure[A captured image.]{\includegraphics[width=0.4\columnwidth]{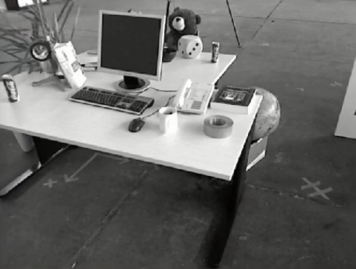}}
	\subfigure[Ground truth depth from Microsoft Kinect.]{\includegraphics[width=0.4\columnwidth]{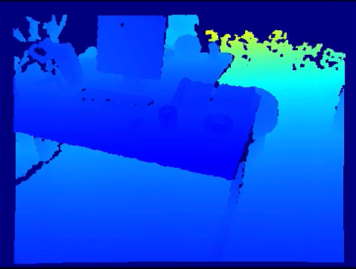}}
	\subfigure[Depth from our approach (T+S+D+H).]{\includegraphics[width=0.4\columnwidth]{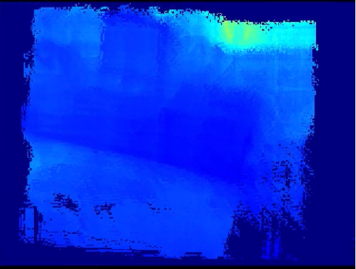}}
	\subfigure[Depth from REMODE \cite{REMODE}.]{\includegraphics[width=0.4\columnwidth]{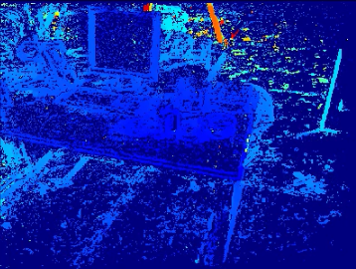}}
	\subfigure[Depth from VI-MEAN \cite{vi_mean}.]{\includegraphics[width=0.4\columnwidth]{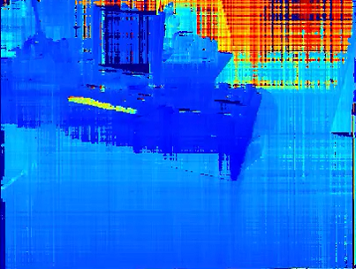}}
	\caption[A visual comparison between our proposed approach and state-of-the-art methods at one of frames in the freiburg2\_desk testing sequence.]{A visual comparison between our proposed approach and state-of-the-art methods (REMODE \cite{REMODE} and VI-MEAN \cite{vi_mean}) at one of frames in the freiburg2\_desk testing sequence. (a) A captured image. (b) Corresponding ground truth depth from Microsoft Kinect. (c) Depth estimation from our approach (T+S+D+H). (d) Depth estimation from REMODE \cite{REMODE}. (e) Depth estimation from VI-MEAN \cite{vi_mean}. Colors vary w.r.t. the distances to the camera. Pixels in dark blue mean no depth estimates.} 
	\label{fig:visual_comparison.}
\end{figure*}

The comparison of average mapping density is shown in Table II. For REMODE \cite{REMODE}, only converged depth estimates are used in the evaluation. Others (i.e. not converged) are not used since they are highly unreliable. Using these depth estimates leads to very low average mapping accuracy. Fig.~\ref{fig:mapping_accuracy_tum.} and Fig.~\ref{fig:mapping_accuracy_icl.} show detailed illustrations of the mapping accuracy on both datasets. We also give a visual comparison between different methods in Fig.~\ref{fig:visual_comparison.} using snapshots from the depth estimations at one of the frames in the freiburg2\_desk testing sequence of the TUM RGB-D dataset. 

We firstly analyze the influence of different components on our approach. The step of temporal cost aggregation is the most time-consuming step. It forms the basis of all following calculations. Applying winter-takes-all strategy after temporal cost aggregation achieves more than 60 average mapping density. However, the corresponding mapping accuracy is very low. The step of spatial regulation, which utilizes the spatial correlation of neighboring depth estimates, not only increases the mapping density, but also increases the mapping accuracy. The local region discussion step slightly reduce the mapping density by rejecting unreliable depth estimates, while the depth refinement step slightly increase the mapping accuracy. The last step, hypothesis filtering, improves the mapping accuracy greatly at the cost of some mapping density reduction. Our hypothesis filtering strategy explicitly makes use of the temporal and spatial correlations of consecutive depth estimates. Consistent depth values are improved while inconsistent ones are removed.

We then compares our approach (T+S+D+H) against REMODE \cite{REMODE} and VI-MEAN \cite{vi_mean}. REMODE \cite{REMODE} runs fastest, as it estimates pixel depth independently, without taking the spatial correlation into consideration. It outputs depth estimates that are in well-textured regions (Fig.~\ref{fig:visual_comparison.}(d)). VI-MEAN \cite{vi_mean} runs slowest and achieves mapping density usually higher than our approach. The main disadvantage of VI-MEAN \cite{vi_mean} is that it does not model outliers, which is demonstrated by the noticeable outliers in its depth estimation (Fig.~\ref{fig:visual_comparison.}(e)). Our approach achieves a good balance between mapping density and mapping accuracy. 
\subsection{Online Indoor and Outdoor Dense Reconstructions}
\begin{figure}[!h]
	\centering
	\subfigure[An indoor image.]{\includegraphics[width=0.48\columnwidth]{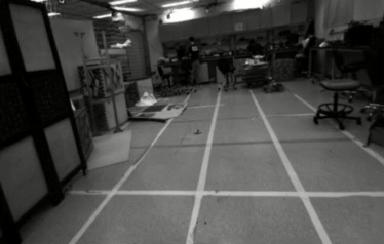}}
	\subfigure[Estimated depth of image (a).]{\includegraphics[width=0.48\columnwidth]{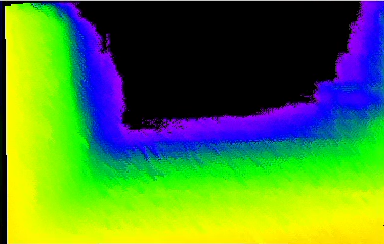}}
	\subfigure[An outdoor image.]{\includegraphics[width=0.48\columnwidth]{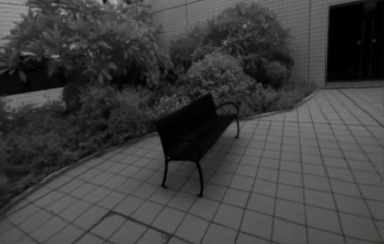}}
	\subfigure[Estimated depth of image (c).]{\includegraphics[width=0.48\columnwidth]{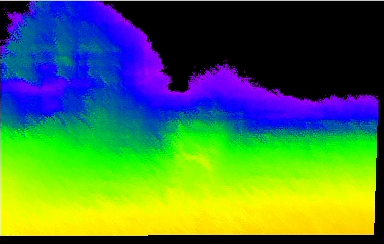}}
	\caption[Snapshots of depths obtained after hypothesis filtering during indoor and outdoor experiments.]{Snapshots of depths obtained after hypothesis filtering during indoor and outdoor experiments. (a)(c) are indoor and outdoor image and (c) (d) are their estimated depths. Colors vary w.r.t. distances to the camera. Dense reconstruction results can be found in Fig.~\ref{fig:exp_indoor.}.}
	\label{fig:snapshots.}
\end{figure}
%


We present online dense reconstructions with a monocular visual-inertial sensor suite. We use the method in \cite{zhenfei_tase} for online pose estimation. The voxel size in the depth integration step is 0.1 meters. Average computation times of depth estimation (T+S+D+H) and uncertainty-aware depth integration are 55.14~ms and 31.18~ms respectively. Snapshots of depths obtained after hypothesis filtering in both indoor and outdoor environments are shown in Fig.~\ref{fig:snapshots.}, while final dense reconstructions are shown in Fig.~\ref{fig:exp_indoor.}. More details of the online depth estimations and dense reconstructions are available at \url{https://1drv.ms/v/s!ApzRxvwAxXqQmlW9ZOrp9hdA7ude}.

\section{Conclusion and Future Work}
\label{sec:conclusions}
In this work, we present a probabilistic approach for monocular dense reconstruction in real time, which makes use of both the spatial and temporal correlations between consecutive depth estimations. In addition to the depth mean, we evaluate its confidence and inlier probability expectation simultaneously in a recursive and probabilistic way. We also take the uncertainty of the depth estimations into account in the depth integration step. Extensive experiments on the TUM RGB-D SLAM dataset and the ICL-NUIM dataset as well as online indoor and outdoor environments demonstrate the effectiveness and efficiency of our presented approach. In the future, we will apply our approach to real-world applications, such as autonomous navigation and AR.

\bibliographystyle{IEEEtran}


\begin{thebibliography}{10}
\providecommand{\url}[1]{#1}
\csname url@rmstyle\endcsname
\providecommand{\newblock}{\relax}
\providecommand{\bibinfo}[2]{#2}
\providecommand\BIBentrySTDinterwordspacing{\spaceskip=0pt\relax}
\providecommand\BIBentryALTinterwordstretchfactor{4}
\providecommand\BIBentryALTinterwordspacing{\spaceskip=\fontdimen2\font plus
\BIBentryALTinterwordstretchfactor\fontdimen3\font minus
  \fontdimen4\font\relax}
\providecommand\BIBforeignlanguage[2]{{%
\expandafter\ifx\csname l@#1\endcsname\relax
\typeout{** WARNING: IEEEtran.bst: No hyphenation pattern has been}%
\typeout{** loaded for the language `#1'. Using the pattern for}%
\typeout{** the default language instead.}%
\else
\language=\csname l@#1\endcsname
\fi
#2}}

\bibitem{SheMicKum1505}
S.~Shen, N.~Michael, and V.~Kumar, ``{Tightly-coupled monocular visual-inertial
  fusion for autonomous flight of rotorcraft {MAV}s},'' in \emph{{Proc. of the
  {IEEE} Intl. Conf. on Robot. and Autom.}}, 2015.

\bibitem{HesKotBow1402}
J.~A. Hesch, D.~G. Kottas, S.~L. Bowman, and S.~I. Roumeliotis, ``{Consistency
  analysis and improvement of vision-aided inertial navigation},''
  \emph{{{IEEE} Trans. Robot.}}, vol.~30, no.~1, pp. 158--176, Feb. 2014.

\bibitem{LiMou1305}
M.~Li and A.~Mourikis, ``{High-precision, consistent {EKF}-based
  visual-inertial odometry},'' \emph{{Intl. J. Robot. Research}}, vol.~32,
  no.~6, pp. 690--711, May 2013.

\bibitem{orb-slam}
R.~Mur-Artal, J.~M.~M. Montiel, and J.~D. Tard\'os, ``{ORB-SLAM}: a versatile
  and accurate monocular {SLAM} system,'' \emph{IEEE Transactions on Robotics},
  vol.~31, no.~5, pp. 1147--1163, 2015.

\bibitem{forster15}
C.~Forster, L.~Carlone, F.~Dellaert, and D.~Scaramuzza, ``{IMU} preintegration
  on manifold for efficient visual-inertial maximum-a-posteriori estimation,''
  in \emph{{Proc. of Robot.: Sci. and Syst.}}, 2015.

\bibitem{NewcombeLD11}
R.~A. Newcombe, S.~Lovegrove, and A.~J. Davison, ``{{DTAM:} Dense tracking and
  mapping in real-time},'' in \emph{{{IEEE} International Conference on
  Computer Vision}}, 2011, pp. 2320--2327.

\bibitem{REMODE}
M.~Pizzoli, C.~Forster, and D.~Scaramuzza, ``{REMODE: Probabilistic}, monocular
  dense reconstruction in real time,'' in \emph{{Proc. of the {IEEE} Intl.
  Conf. on Robot. and Autom.}}, 2014.

\bibitem{MonoFusion}
V.~Pradeep, C.~Rhemann, and S.~Izadi, ``{MonoFusion}: Real-time {3D}
  reconstruction of small scenes with a single web camera,'' in \emph{{IEEE
  International Symposium on Mixed and Augmented Reality}}, 2013.

\bibitem{alpha_beta}
G.~Vogiatzis and C.~Hernandez, ``Video-based, real-time multi-view stereo,''
  \emph{Image and Vision Computing}, vol.~29, pp. 434--441, 2011.

\bibitem{sgm07}
H.~Hirschmuller, ``Stereo processing by semiglobal matching and mutual
  information,'' \emph{IEEE Transactions on Pattern Analysis and Machine
  Intelligence}, vol.~30, no.~2, 2008.

\bibitem{Geiger2010ACCV}
A.~Geiger, M.~Roser, and R.~Urtasun, ``Efficient large-scale stereo matching,''
  in \emph{Asian Conference on Computer Vision}, 2010.

\bibitem{marching_cubes}
W.~E. Lorensen and H.~E. Cline, ``Marching cubes: A high resolution {3D}
  surface construction algorithm,'' \emph{SIGGRAPH Comput. Graph.}, vol.~21,
  no.~4, pp. 163--169, Aug. 1987.

\bibitem{kinectfusion}
N.~Richard, I.~Shahram, H.~Otmar, M.~David, K.~David, D.~Andrew, K.~Pushmeet,
  S.~Jamie, H.~Steve, and F.~Andrew, ``Kinectfusion: Real-time dense surface
  mapping and tracking,'' in \emph{The {IEEE} International Symposium on Mixed
  and Augmented Reality}, October 2011.

\bibitem{Niebner2013}
M.~Nie{\ss}ner, M.~Zollh{\"{o}}fer, S.~Izadi, and M.~Stamminger, ``Real-time
  {3D} reconstruction at scale using voxel hashing,'' \emph{ACM Trans. Graph.},
  vol.~32, no.~6, 2013.

\bibitem{Whelan16ijrr}
T.~Whelan, R.~F. Salas-Moreno, B.~Glocker, A.~J. Davison, and S.~Leutenegger,
  ``Elasticfusion: Real-time dense {SLAM} and light source estimation,''
  \emph{Intl. J. of Robotics Research}, 2016.

\bibitem{vi_mean}
Y.~Lin, F.~Gao, T.~Qin, W.~Gao, T.~Liu, W.~Wu, Z.~Yang, and S.~Shen,
  ``Autonomous aerial navigation using monocular visual-inertial fusion,''
  \emph{Journal of Field Robotics}, 2017.

\bibitem{Stuhmer2010}
J.~St{\"u}hmer, S.~Gumhold, and D.~Cremers, ``Real-time dense geometry from a
  handheld camera,'' in \emph{Proceedings of the DAGM Symposium on Pattern
  Recognition}, 2010.

\bibitem{3DModelingOnTheGO}
T.~Schoeps, T.~Sattler, C.~Hane, and M.~Pollefeys, ``{3D} modeling on the go:
  Interactive {3D} reconstruction of large-scale scenes on mobile devices,'' in
  \emph{Proceedings of International Conference on 3D Vision}, 2015.

\bibitem{multi_level_mapping}
W.~N. Greene, K.~Ok, and P.~Lommel, ``Multi-level mapping: Real-time dense
  monocular {SLAM},'' in \emph{{Proc. of the {IEEE} Intl. Conf. on Robot. and
  Autom.}}, 2016.

\bibitem{dpptam}
A.~Concha and J.~Civera, ``Dense piecewise planar tracking and mapping from a
  monocular sequence,'' in \emph{{Proc. of the {IEEE/RSJ} Intl. Conf. on
  Intell. Robots and Syst.}}, 2015.

\bibitem{superpixel_expansion}
L.~Teixeira and M.~Chli, ``Real-time local {3D} reconstruction for aerial
  inspection using superpixel expansion,'' in \emph{{Proc. of the {IEEE} Intl.
  Conf. on Robot. and Autom.}}, 2017.

\bibitem{zhenfei_tase}
{Z. Yang and S. Shen}, ``{Monocular visual-inertial state estimation with
  online initialization and camera-{IMU} extrinsic calibration},'' \emph{IEEE
  Transactions on Automation Science and Engineering}, vol.~14, pp. 39--51,
  2017.

\bibitem{BCML96}
B.~Curless and M.~Levoy, ``A volumetric method for building complex models from
  range images,'' in \emph{Proceedings of the 23rd Annual Conference on
  Computer Graphics and Interactive Techniques}, 1996.

\bibitem{Klingensmith2015}
M.~Klingensmith, I.~Dryanovski, S.~Srinivasa, and J.~Xiao, ``{CHISEL} : Real
  time large scale {3D} reconstruction onboard a mobile device using
  spatially-hashed signed distance fields,'' in \emph{{Proc. of Robot.: Sci.
  and Syst.}}, 2015.

\bibitem{CHOMP}
N.~Ratliff, M.~Zucker, J.~Bagnell, and S.~Srinivasa, ``{CHOMP}: Gradient
  optimization techniques for efficient motion planning,'' in \emph{{Proc. of
  the {IEEE} Intl. Conf. on Robot. and Autom.}}, May 2009.

\bibitem{HelenOleynikova16}
H.~Oleynikova, M.~Burri, Z.~Taylor, J.~Nieto, R.~Siegwart, and E.~Galceran,
  ``Continuous-time trajectory optimization for online {UAV} replanning,'' in
  \emph{{Proc. of the {IEEE/RSJ} Intl. Conf. on Intell. Robots and Syst.}}, Oct
  2016.

\end{thebibliography}

\end{document}